\documentclass[runningheads]{llncs}
\usepackage{graphicx}
\usepackage{comment}
\usepackage{amsmath,amssymb} % define this before the line numbering.
\usepackage{color}

\usepackage{mmstyle}
\usepackage{booktabs} % To thicken table lines

\usepackage{algorithm}
\usepackage{algorithmicx}
\usepackage{algpseudocode}
\usepackage{enumitem}
\usepackage{dsfont}

\usepackage[font=small,labelfont=bf]{caption}
\begin{document}
\pagestyle{headings}
\mainmatter
\def\ECCVSubNumber{2376}  % Insert your submission number here

\title{Learn to Propagate Reliably on \\ Noisy Affinity Graphs}
% Replace with your title

% INITIAL SUBMISSION 
\begin{comment}
\titlerunning{ECCV-20 submission ID \ECCVSubNumber} 
\authorrunning{ECCV-20 submission ID \ECCVSubNumber} 
\author{Anonymous ECCV submission}
\institute{Paper ID \ECCVSubNumber}
\end{comment}
%******************

% CAMERA READY SUBMISSION
%\begin{comment}
\titlerunning{Learn to Propagate Reliably on Noisy Affinity Graphs}
% If the paper title is too long for the running head, you can set
% an abbreviated paper title here
%
\author{Lei Yang\inst{1}\orcidID{0000-0002-0571-5924} \and
Qingqiu Huang\inst{1}\orcidID{0000-0002-6467-1634} \and \\
Huaiyi Huang\inst{1}\orcidID{0000-0003-1548-2498} \and
Linning Xu\inst{2}\orcidID{0000-0003-1026-2410} \and
Dahua Lin\inst{1}\orcidID{0000-0002-8865-7896}}
\authorrunning{L. Yang and Q. Huang and H. Huang and L. Xu and D. Lin}
% First names are abbreviated in the running head.
% If there are more than two authors, 'et al.' is used.
%
\institute{The Chinese University of Hong Kong \and
The Chinese University of Hong Kong, Shenzhen \\
%\email{lncs@springer.com}\\
%\url{http://www.springer.com/gp/computer-science/lncs} \and
%ABC Institute, Rupert-Karls-University Heidelberg, Heidelberg, Germany\\
\email{\{yl016,hq016,hh016,dhlin\}@ie.cuhk.edu.hk,linningxu@link.cuhk.edu.cn}}
%\end{comment}
%******************
\maketitle

%%%%%%%%% BODY TEXT
% !TEX root = ../main.tex

\begin{abstract}
Recent works have shown that exploiting unlabeled data through label
propagation can substantially reduce the labeling cost, which has been
a critical issue in developing visual recognition models.
Yet, how to propagate labels reliably,
especially on a dataset with unknown outliers,
remains an open question.
Conventional methods such as linear diffusion lack the capability of handling
complex graph structures and may perform poorly when the seeds are sparse.
Latest methods based on graph neural networks would face difficulties
on performance drop as they scale out to noisy graphs.
To overcome these difficulties, we propose a new framework that allows
labels to be propagated reliably on large-scale real-world data.
This framework incorporates (1) a local graph neural network to predict
accurately on varying local structures while maintaining high scalability,
and (2) a confidence-based path scheduler that identifies outliers and
moves forward the propagation frontier in a prudent way.
%and (3) a confidence estimator to identify outliers to ensure the reliability of propagation.
Both components are learnable and closely coupled.
Experiments on both ImageNet and Ms-Celeb-1M show that our confidence guided framework can significantly improve the overall accuracies of the propagated labels,
especially when the graph is very noisy.
\end{abstract}

% !TEX root = ../main.tex

\section{Introduction}
\label{sec:introduction}
\begin{figure}[t]
	\centering
	\includegraphics[width=\linewidth]{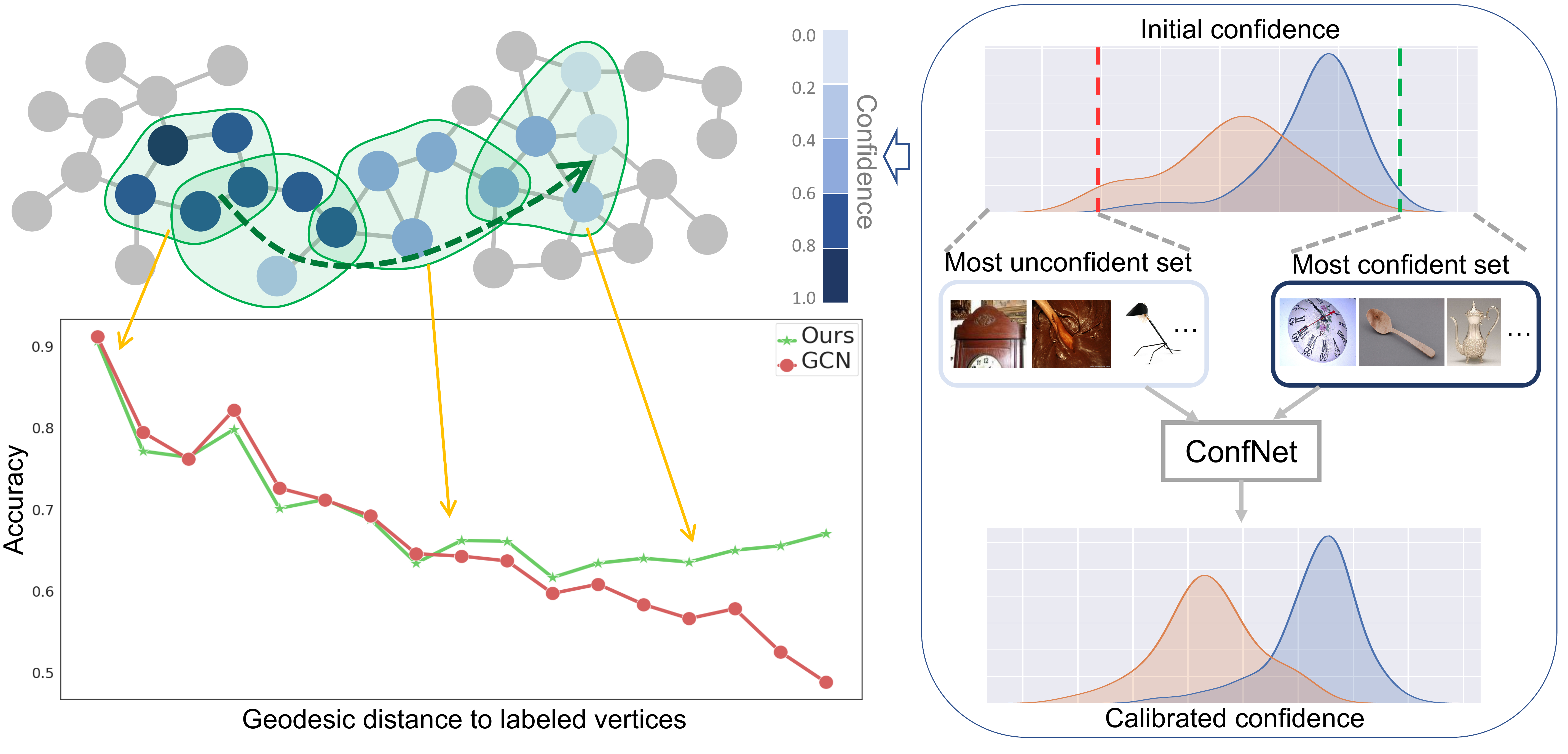}
	\caption{
In this paper, we propose a framework for transductive learning on noisy graphs,
which contain a large number of outliers, \eg out-of-class samples.
The framework consists of a local predictor and a confidence-based path scheduler.
The predictor updates local patches sequentially following a path driven by the estimated confidences.
The path scheduler leverages both the confident and unconfident samples from the predictor to further calibrate the estimated confidence.
The unconfident samples are usually images with low quality(\eg the leftmost image is a clock with only top part),
hard examples(\eg the middle image is a spoon mixed with the background)
or \emph{out-of-class} samples(\eg the rightmost image is a lamp but none of the labeled samples belong to this class).
The lower left figure experimentally shows that the proposed method improves the reliability of propagation.
When the distance from unlabeled samples to labeled ones increases,
our method surpasses state-of-the-art by a significant margin
	}
	\label{fig:teaser}
\end{figure}

% Importance of the problem
%
The remarkable advances in visual recognition~\cite{deng2009imagenet,sun2014deep,schroff2015facenet,he2016deep,hu2018squeeze,zhang2018accelerated,deng2018arcface,xia2020online,huang2020placepedia,huang2018trailers,Huang_2018_CVPR,Xiong_2019_ICCV,rao2020local,xia2020online,huang2020movie,rao2020unified,wu2020distribalanced}
are built on top of large-scale annotated training data.
However, the ever increasing demand on annotated data has resulted in prohibitive annotation cost.
%
%Semi-supervised learning, which aims to
%leverage unlabeled samples in conjunction with the labeled ones,
Transductive learning, which aims to
propagate labeled information to unlabeled samples,
is a promising way to tackle this issue.
Recent studies
~\cite{zhu2005semi,perozzi2014deepwalk,kipf2016semi,velivckovic2017graph,oliver2018realistic,huang2020caption}
show that transductive methods with an appropriate
design can infer unknown labels accurately while dramatically reducing the
annotation efforts.

% Summarize previous methods and limitations
%
Many transductive methods adopt graph-based propagation
~\cite{zhu2002learning,perozzi2014deepwalk,kipf2016semi,velivckovic2017graph}
as a core component.
Generally, these methods construct a graph among all samples, propagating
labels or other relevant information from labeled samples to unlabeled ones.
Early methods~\cite{zhu2002learning,zhou2004learning,belkin2006manifold}
often resort to a linear diffusion paradigm, where the class probabilities
for each unlabeled sample are predicted as a linear combination of those for its neighbors.
%Despite the simplicity, such methods lack the capability of dealing with complicated graph structures.
Relying on simplistic assumptions restricts their capability of dealing with complicated graph structures in real-world datasets.
%Despite the simplicity, such methods are sensitive to noisy predictions -- a
%wrong prediction can severely mislead those downstream.
%This problem gets worse on massive data with complicated graphical structures.
%
%Recently, there have been attempts to exploit the strong expressive power
%of graph convolutional network~\cite{kipf2016semi,velivckovic2017graph,wu2019simplifying}
%and obtain encouraging results.
Recently, graph convolutional networks~\cite{kipf2016semi,velivckovic2017graph,wu2019simplifying}
have revealed its strong expressive power to process complex graph structures.
Despite obtaining encouraging results, these GCN-based methods remain limited in an important aspect,
namely the capability of coping with outliers in the graph.
In real-world applications, unlabeled samples do not necessarily share
the same classes with the labeled ones, leading to a large portion of
\emph{out-of-class} samples, which becomes the main source of outliers.
%Besides, the image quality and appearance variance also makes the real-world dataset very noisy.
%These two types of data constitute a large number of outliers in the unlabeled data.
Existing methods ignore the fact that the confidences
of predictions on different samples can vary significantly,
which may adversely influence the reliability of the predictions.
%First, they operate layer-by-layer on the entire graph, which would lead
%to prohibitive demand on memory when scaled out to massive graphs,
%\eg~those with millions of vertices.
%More importantly, existing methods ignore the fact that the confidences
%of the predictions on different samples can vary significantly,
%which may adversely influence the reliability of the predictions,
%especially in the real-world graph with unknown noise.
%\todo{namely, they do not necessarily belong to the same class with the labeled seeds.}
%Second, the propagation takes place in parallel across all vertices,
%regardless of the fact that the confidences of the predictions on different
%samples
%can vary significantly, which may adversely influence the reliability of the
%predictions.

% overview of our method
In this paper, we aim to explore a new framework that can propagate labels
over noisy unlabeled data \emph{reliably}.
This framework is designed based on three principles:
1) \emph{Local update:} each updating step can be carried out within
a local part of the graph, such that the algorithm can be easily scaled out
to a large-scale graph with millions of vertices.
% without incurring excessive memory demand.
2) \emph{Learnable:} the graph structures over a real-world dataset
are complex, and thus it is difficult to prescribe a rule that works well
for all cases, especially for various unknown outliers.
Hence, it is desirable to have a core operator with strong
expressive power that can be learned from real data.
3) \emph{Reliable path:} graph-based propagation is sensitive to noises --
a noisy prediction can mislead other predictions downstream.
To propagate reliably, it is crucial to choose a path such that most inferences
are based on reliable sources.

%\todo{address both local predictor and confidence estimator are learnable}
Specifically, we propose a framework comprised of two learnable components,
namely, a local predictor and a path scheduler.
The local predictor is a light-weight graph neural network operating on
local sub-graphs, which we refer to as \emph{graph patches}, to predict
the labels of unknown vertices.
The path scheduler is driven by confidence estimates, ensuring that labels
are gradually propagated from highly confident parts to the rest.
The key challenge in designing the path scheduler is 
how to estimate the confidences effectively.
We tackle this problem via a two-stage design.
First, we adopt a \emph{multi-view} strategy by exploiting the fact that a vertex is usually covered by multiple \emph{graph patches},
where each patch may project a different prediction on it.
Then the confidence can be evaluated on how consistent and certain the
predictions are.
Second, with the estimated confidence, we construct a candidate set by
selecting the most confident samples and the most unconfident ones.
As illustrated in Fig.~\ref{fig:teaser}, we devise a \emph{ConfNet}
to learn from the candidate set and calibrate the confidence
estimated from the first stage.
%As illustrated in Figure.~\ref{fig:teaser}, the \emph{ConfNet} enlarges the gap between noisy unlabeled samples and normal unlabeled samples.
%
%We tackle this problem using a \emph{multi-view} strategy by exploiting
%the fact that a vertex is usually covered by multiple \emph{graph patches},
%where each patch may project a different prediction on it.
%Then the confidence can be evaluated on how consistent and certain the
%predictions are.
Highly confident samples are assumed to be labeled and used
in later propagation, while highly unconfident samples are assumed to
be outliers and excluded in later propagation.
Both components work closely together to drive the propagation process.
On one hand, the local predictor follows the scheduled path to update
predictions; on the other hand, the path scheduler estimates confidences
based on local predictions.
Note that the training algorithm also follows the same coupled procedure,
where the parameters of the local predictor and confidence estimator are learned end-to-end.
%\todo{address that it is hard to use supervised method for noise prediction, as noise may vary a lot, in this sense, transductive approach is very suitable for this}

Our main contributions lie in three aspects:
(1) A learnable framework that involves a local predictor and a path scheduler
to drive propagation reliably on noisy large-scale graphs.
(2) A novel scheme of exploiting both confident and unconfident samples for confidence estimation.
(3) Experiments on ImageNet~\cite{deng2009imagenet} and
Ms-Celeb-1M~\cite{guo2016ms} show that our proposed approach
outperforms previous algorithms, especially when the graphs are noisy and the initial seeds are sparse.
%With only $1\%$ of labeled samples,
%the algorithm raises the inference accuracy
%from $49.11\%$ to $52.07\%$ on ImageNet,
%and from $89.06\%$ to $92.15\%$ on Ms-Celeb-1M,
%comparing to the previous state-of-the-art.

% !TEX root = ../main.tex

\section{Related Work}
\label{sec:related}

%Semi-supervised Learning aims to exploit large amounts of unlabeled
%data with a limited number of labeled data.
%There are two settings in semi-supervised learning, namely \emph{transductive learning} and \emph{inductive learning}.
%%
%Given a labeled set and an unlabeled set, the former aims to perform
%predictions only for the test samples, while the latter one tries to
%output a prediction function over the entire space.
%
In this paper, we focus on graph-based transductive learning
~\cite{zhu2002learning,perozzi2014deepwalk,kipf2016semi,velivckovic2017graph,huang2018person},
which constructs a graph among all samples and
propagates information from labeled samples to unlabeled ones.
We summarize existing methods into three categories and briefly
introduce other relevant techniques.

\noindent \textbf{Early Methods.}
% conventional methods
Conventional graph-based transductive learning~\cite{zhu2002learning,zhou2004learning,belkin2006manifold}
is mainly originated from smoothness assumption, which is formulated as a graph Laplacian regularization.
%Label propagation~\cite{zhu2002learning}, label spreading~\cite{zhou2004learning},
%and manifold regularization~\cite{belkin2006manifold} belong to this category.
They share the same paradigm to aggregate neighbors' information through linear combination.
While relying on the simple assumption, these methods are limited
by their capability of coping with complex graph structures in large-scale real-world datasets.
%, especially when applied to graphs with massive vertices.

% GCN
\noindent \textbf{GCN-based Methods.}
Graph Convolutional Network (GCN) and its
variants~\cite{kipf2016semi,velivckovic2017graph,schlichtkrull2018modeling,wu2019simplifying}
apply filters over the entire graph and achieve impressive performance on several tasks~\cite{zhou2018graph}.
To extend the power of GCN to large-scale graphs, 
GraphSAGE~\cite{hamilton2017inductive} proposes to
sample a fixed number of neighbors and apply aggregation functions thereon.
FastGCN~\cite{chen2018fastgcn} further reduces memory demand by
sampling vertices rather than neighbors in each graph convolution layer.
%
%Graph Partition Neural Network (GPNN)~\cite{liao2018graph} alternates between local
%propagation among nodes in each subgraph and global propagation among these subgraphs.
% SGC
%
However, they propagate labels in parallel across all vertices,
regardless of the confidence difference among predictions.
This ignorance on prediction confidence may adversely influence
the reliability of propagation.

% Confidence-based methods
\noindent \textbf{Confidence-based Methods.}
Previous approaches either model the node label as a distribution along
with uncertainty~\cite{saunders1999transduction,bojchevski2017deep}
or re-scale the weight of links by introducing attention to vertices~\cite{velivckovic2017graph,vashishth2019confidence}.
Unlike these methods, which mainly focus on making use of confident samples,
our approach learns from both confident and unconfident data for confidence estimation.
%Unlike these methods, which require multiple models or introduce extra parameters
%to formulate confidence, we estimate the confidence by observing the data from different subgraphs under the same model, which is more efficient.
%
%The estimated confidence is then used to drive propagation,
%which leads to a more reliable propagation scheme.

% inductive learning
\noindent \textbf{Inductive Learning.}
Inductive learning is closely related to transductive learning~\cite{vapnik200624}.
The key difference lies in that the former aims to learn a better
model with unlabeled data~\cite{tarvainen2017mean,oliver2018realistic,zhan2018consensus,yang2019learning,yang2020learning},
while the latter focuses on predicting labels for unlabeled
data~\cite{zhu2002learning,zhu2005semi}.
One important line of inductive learning is leveraging the predicted labels
of unlabeled data in a supervised manner~\cite{lee2013pseudo,iscen2019label}.
In this paper, we focus on transductive learning and show that the obtained
pseudo labels can be applied to inductive learning as a downstream task.

% universum
%\noindent \textbf{Universum Learning.}
%%
%Universum~\cite{weston2006inference} assumes that out-of-class samples
%are available for training and have been shown to benefit supervised learning via a regularization effect~\cite{chapelle2008analysis,zhang2017universum}.
%%
%However, given a set of unlabeled data in real practice, they do not
%necessarily belong to the same classes with the labeled seeds and it is
%hard to know out-of-class data in advance.
%In this paper, we focus on this practical setting and regard out-of-class samples as outliers.
%\textcolor{blue}{In this paper, we focus on this practical setting and regard out-of-class samples as outliers}.

% anomaly detection
\noindent \textbf{Outlier Detection.}
Previous methods~\cite{chandola2009anomaly} either rely on crafted unsupervised rules~\cite{noble2003graph}
or employing a supervised method to learn from an extra labeled outlier dataset~\cite{cheng2016wide}.
The unsupervised rules lack the capability of handling complex real-world dataset,
while the supervised methods are easy to overfit to the labeled outliers and do not generalize well.
In this paper, we \emph{learn} to identify outliers from carefully
selected confident and unconfident samples during propagation.
% !TEX root = ../main.tex

% Methodology

\section{Propagation on Noisy Affinity Graphs}

Our goal is to develop an effective method to
propagate reliably over noisy affinity graphs, \eg~those containing lots of out-of-class samples, while maintaining reasonable runtime cost.
This is challenging especially when the proportion of outliers are high and initial seeds are sparse.
As real-world graphs often have complex and varying structures,
noisy predictions can adversely affect these downstream along the
propagation paths.
We propose a novel framework for graph-based propagation, which
copes with the complexity in local graph structures via a light-weight
graph convolutional network while improving the reliability via a
confidence-based scheduler that chooses propagation paths prudently.

\begin{figure*}[t]
	\centering
	\includegraphics[width=0.9\linewidth]{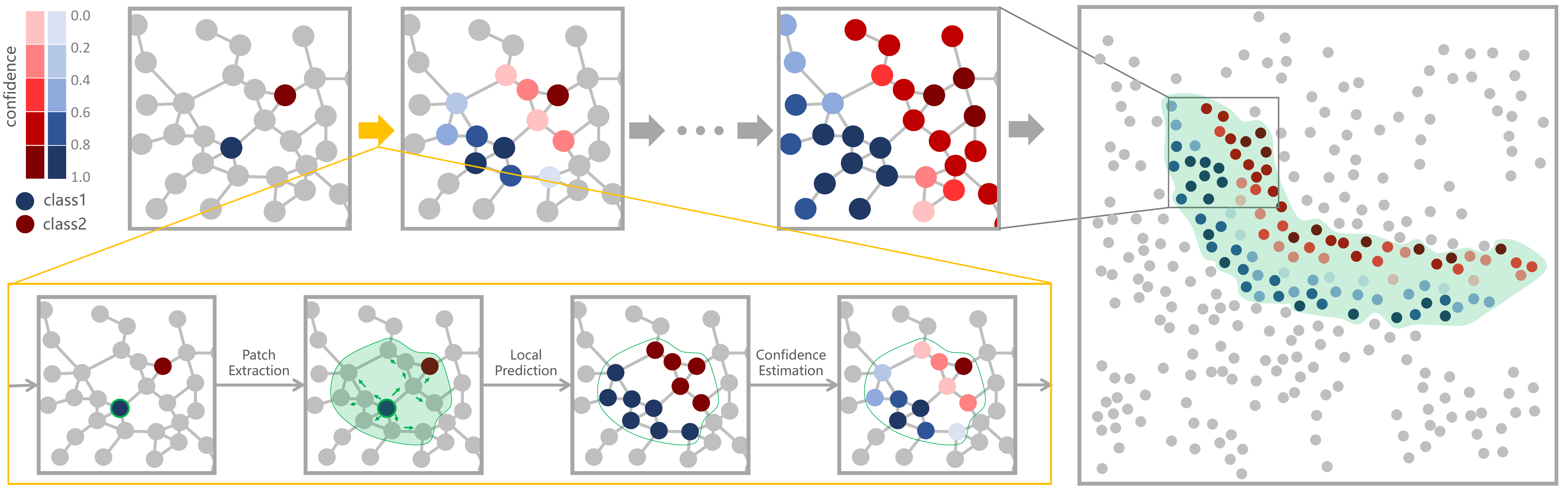}
	\caption{\small
		Overview of our framework (better viewed in color).
		At each iteration, our approach consists of three steps:
		(1) Starting from the selected confident vertex, the patch extractor
		generates a graph patch.
		(2) Given the graph patch, the learned local predictor updates the
		predictions of all unlabeled vertices on the patch.
		(3) Given the updated predictions, the path scheduler estimates
		confidence for all unlabeled vertices.
		Over many iterations, labeled information are gradually
		propagated from highly confident parts to the rest
	}
	\label{fig:framework}
\end{figure*}

\subsection{Problem Statement}

Consider a dataset with $N = N_l + N_u$ samples, where $N_l$ samples are labeled and
$N_u$ are unlabeled, and $N_l \ll N_u$.
We denote the set of labeled samples as $\cD_l = \{(\vx_i, y_i)\}^{N_l}_{i=1}$,
and that of unlabeled ones as $\cD_u = \{\vx_i\}^{N_l+N_u}_{i=N_l+1}$.
Here, $\vx_i \in \mathbb{R}^d$ is the feature for the $i$-th sample,
which is often derived from a deep network in vision tasks,
and $y_i \in \cY$ is its label, where $\cY = \{1, . . . , m\}$.
In our setting, $\cD_u$ consists of two parts, namely in-class samples and out-of-class samples.
For out-of-class data, their labels do not belong to $\cY$.
The labeled set $\cD_l$ only contains in-class labeled samples.
The goal is to assign a label $\hat{y} \in \cY\bigcup\{-1\}$ to each unlabeled sample in $\cD_u$,
where $\hat{y}=-1$ indicates an unlabeled sample is identified as an outlier.

% graph definition
To construct an affinity graph $\cG = (\cV, \cE)$ on this dataset,
we treat each sample as a vertex and connect it with its $K$ nearest neighbors.
The graph $\cG$ can be expressed by an adjacency matrix
$\mA \in \mathbb{R}^{N \times N}$,
where $a_{i, j} = \mA(i, j)$ is the cosine similarity
between $\vx_i$ and $\vx_j$ if $(i, j) \in \cE$, and otherwise $0$.

For label propagation, we associate each vertex with a
probability vector $\vp_i$, where $p_{ik} = \vp_i(k)$ indicates
the probability of the sample $\vx_i$ belonging to the $k$-th class,
and a confidence score $c_i \in [0, 1]$.
For labeled samples, $\vp_i$ is fixed to be a one-hot vector
with $p_{ik} = 1$ for $k = y_i$.
For unlabeled samples, $\vp_i$ is initialized to be a uniform
distribution over all classes and will be gradually updated as the propagation
proceeds.
We set a threshold $o_{\tau}$ to determine whether a sample is an outlier.
After the propagation is completed,
for those with confidence smaller than $o_{\tau}$, the predicted label for each unlabeled sample $\vx_i$ is set to $\hat{y}_i=-1$.
For those with confidence larger than $o_{\tau}$.
the predicted label for each
unlabeled sample $\vx_i$ is set to be the class with highest probability
in $\vp_i$, as $\hat{y}_i = \argmax_k p_{ik}$.

\subsection{Algorithm Overview}

As shown in Fig.~\ref{fig:framework}, our proposed propagation scheme is based
on graph patches as the units for updating.
Here, a \emph{graph patch} is a sub-graph containing both labeled and
unlabeled vertices. The algorithm performs updates over a graph patch in each
step of propagation.

The propagation proceeds as follows.
(1) At each iteration, we first randomly select a vertex from the
\emph{high-confidence vertex set} $\cS$, which contains both the
initially labeled samples and those samples whose confidences are high enough
to be considered as ``labeled'' as the propagation proceeds.
%$\cS$ is initially set to contain only the labeled samples and will
%grow over time.
%
(2) Starting from the selected \emph{confident vertex}, we use a patch extractor
to expand it into a graph patch, and then update the predictions on all
unlabeled vertices in this patch,
%\ie~those not in $\cS$,
using a local predictor.
(3) The path scheduler uses these predictions to re-estimate
confidences for unlabeled vertices.
In this work, both the local predictor and the path scheduler are
formulated as a graph convolutional network (GCN) learned from the training data,
in order to cope with the complexity of local graph structures.
All the vertices whose confidence scores
go beyond a threshold $c_\tau$ will be added into $\cS$ and their
predictions will not be updated again in future iterations.
Note that the updated confidences would influence the choice
of the next confident vertex and thus the propagation path.
By iteratively updating predictions and confidences as above,
the algorithm drives the propagation across
the entire graph, gradually from high confident areas to the rest.

This propagation algorithm involves two components:
%a \emph{patch extractor} for generating a graph patch from a confident vertex,
a \emph{local predictor} that generates confident graph patches and updates predictions thereon, and
a \emph{path scheduler} that estimates confidences and schedules the propagation
path accordingly.
Next, we will elaborate on these components in turn.

%\subsection{Patch Extractor}
%\label{sec:method_patch}

\subsection{GCN-based Local Predictor}
\label{sec:method_pred}

\noindent \textbf{Patch extractor.}
A graph patch with the following properties is a good candidate for the next
update.
(1) \emph{High confidence:} We define the \emph{confidence of a graph patch}
as the sum of its vertex confidences.
A patch with high confidence is more likely to yield reliable predictions
due to the availability of reliable information sources.
(2) \emph{Large expected confidence gain}:
We define the \emph{estimated confidence gain} of a patch $\cP_i$ as
$\sum_{v_j \in \cP_i} (1 - c_j)$, \ie~the maximum possible improvement
on the total confidence. Performing updates on those patches with
large expected confidence gain can potentially speed up the propagation. To
maintain sufficient confidence gain while avoiding excessive patch sizes, we
consider a patch as \emph{viable} for the next update if the expected gain is
above a threshold $\Delta c_\tau$ and the size is below the maximum size $s$.
Besides, to avoid selecting highly overlapped patches, once a vertex
is taken as the start point, its $m$-hop neighbors will all be
excluded from selecting as start points in later propagation.

%Alg.~\ref{alg:patch} shows the detailed procedure to produce a graph patch $\cP$:
To generate a graph patch $\cP$,
we start from the most confident vertex and add its immediate neighbors into a queue.
Each vertex in the queue continues to search its unvisited
neighbors until
(1) the expected gain is above $\Delta c_\tau$, which means that a
viable patch is obtained; or
(2) the size exceeds $s$, which means that no viable patch is found
around the selected vertex and the algorithm randomly selects a new
vertex from $\cS$ to begin with.
Note that our propagation can be parallelized by selecting multiple non-overlapped patches at the same time.
We show the detailed algorithm in supplementary.

Graph patches are dynamically extracted along with the propagation.
In early iterations, $\Delta c_\tau$ can often be achieved by
a small number of unlabeled vertices, as most vertices are unlabeled
and have low confidences. This results in more conservative exploration at the
early stage.
As the propagation proceeds, the number of confident vertices
increases while the average expected confidence gain decreases,
the algorithm encourages more aggressive updates over larger patches.
Empirically, we found that on an affinity graph with $10K$ vertices with
$1\%$ of labeled seeds,
it takes about $100$ iterations to complete the propagation procedure,
where the average size of graph patches is $1K$.
%leave a pointer to sec.5.4?
%We provide a more detailed study on the patch extraction efficiency in
%the supplemental materials.

\noindent \textbf{Design of local predictor.}
We introduce a graph convolutional network (GCN) to predict unknown labels
for each graph patch.
Given a graph patch $\cP_i$ centered at $v_i \in \cV$,
the network takes as input the visual features $\vx_i$,
and the affinity sub-matrix restricted to $\cP_i$, denoted as $\mA(\cP_i)$.
Let $\mF_0(\cP_i)$ be a matrix of all vertex data for $\cP_i$,
where each row represents a vertex feature $\vx_i$.
The GCN takes $\mF_0(\cP_i)$ as the input to the bottom layer and
carries out the computation through $L$ blocks as follows:
\begin{equation}
\mF_{l+1}(\cP_{i}) = \sigma\left(
\tilde{\mD}(\cP_{i})^{-1} \tilde{\mA}(\cP_i) \mF_l(\cP_{i})\mW_l
\right),
\label{eq:gcn}
\end{equation}
where $\tilde{\mA}(\cP_i) = \mA(\cP_i) + \mI$;
$\tilde{\mD} = \sum_j \tilde{\mA}_{ij}(\cP_{i})$ is a diagonal
degree matrix;
$\mF_l(\cP_{i})$ contains the embeddings at the $l$-th layer;
$\bold{W}_{l}$ is a matrix to transform the embeddings;
$\sigma$ is a nonlinear activation (\emph{ReLU} in this work).
Intuitively, this formula expresses a procedure of taking weighted average
of the features of each vertex and its neighbors based on affinity weights,
transforming them into a new space with $\bold{W}_{l}$, and then feeding them
through a nonlinear activation.
Note that this GCN operates locally within a graph patch and thus the demand
on memory would not increase as the whole graph grows, which makes it easy to
scale out to massive graphs with millions of vertices.

As the propagation proceeds, each vertex may be covered
by multiple patches, including those constructed in previous steps.
Each patch that covers a vertex $v$ is called a \emph{view}
of $v$.
We leverage the predictions from multiple views for higher reliability,
and update the probability vector for each unlabeled vertex
in $\cP_i$ by averaging the predictions from all views, as
\begin{equation}
\vp_i = \dfrac{1}{\sum \mathds{1}_{v_i \in \cP_j}} \sum_{v_i \in \cP_j}
\mF_{L}(v_{i,j}).
\label{eq:pred}
\end{equation}

% alg for extracting a single patch
%\begin{algorithm}[t]
%	\caption{Patch Extraction}
%	\begin{algorithmic}[1]
%		\renewcommand{\algorithmicrequire}{\textbf{Input:}}
%		\renewcommand{\algorithmicensure}{\textbf{Output:}}
%		\Require Confidence ranked set $\{(v_i, c_i)\}_{i=1}^N$,
%		max size $s$, minimum expected gain $\Delta c_\tau$
%		\Ensure  Patch $\cP$
%		\State $\cP = \emptyset$, $Q = \{(v_0, c_0)\}$
%		\While{$Q \neq \emptyset$}
%		\State $v$, $c$ = \Call{Pop}{$Q$}
%		\If{$v$ is not visited}
%		\State $\cP = \cP \cup \{(v, c)\}$
%		\If{\Call{ExpectedGain}{$\cP$} $> \Delta c_\tau$ or $|P| > s$}
%		\State break
%		\EndIf
%		\State \Call{AddNeighborToQueue}{$v$, $Q$}
%		\EndIf
%		\EndWhile
%		\State \Return $\cP$
%	\end{algorithmic}
%	\label{alg:patch}
%\end{algorithm}

\subsection{Confidence-based Path Scheduler}
\label{sec:method_prop}

Confidence estimation is the core of the path scheduler.
A good estimation of confidences is crucial for reliable propagation, as
it allows unreliable sources to be suppressed.
%
%The fact that humans can identity samples with unknown classes
%suggests that this goal is achievable in principle.
%
Our confidence estimator involves
a \emph{Multi-view} confidence estimator and a learnable \emph{ConfNet},
to form a two-stage procedure.
Specifically,
the former generates an initial confidence estimation by aggregating
predictions from multiple patches.
Then ConfNet learns from the most confident samples and the most unconfident ones from the first stage, to further refine the confidence.
The ultimate confidence is the average confidence of these two stages.

\noindent \textbf{Multi-view confidence estimation.}
%\todo{add intuition of multi-view: it shares the similar spirit with multiple image patch testing, which can be regarded as a single model ensemble approach.}
%
%Existing methods to estimate the confidences of neural network predictions
%mainly rely on dropout~\cite{gal2016dropout} or
%network ensembles~\cite{lakshminarayanan2017simple}, which are expensive.
%
Previous studies~\cite{gal2016dropout,lakshminarayanan2017simple}
have shown that neural networks usually yield over-confident predictions.
In this work, we develop a simple but effective way to alleviate the over-confidence problem.
We leverage the multiple views for each vertex $v_i$ derived along the propagation process.
Particularly, the confidence for $v_i$ is defined as
\begin{equation}
c_i =
\begin{cases}
\max_k p_{ik}, & \text{if $v_i$ was visited multiple times}\\
\epsilon,      & \text{if $v_i$ was visited only once}
\end{cases}
\label{eq:conf}
\end{equation}
where $\vp_{i}$ is given in Eq.~\eqref{eq:pred},
and $\epsilon$ is a small positive value.

Here, we discuss why we use $c_i$ as defined above to measure
the confidence.
When a vertex has only been visited once, it is difficult to assess
the quality of the prediction, therefore it is safe to assume
a low confidence.
When a vertex has been visited multiple times, a high value of
$\max_k p_{ik}$ suggests that the
predictions from different views are consistent with each other.
If not, \ie~different views vote for different
classes, then the average probability for the best class would be
significantly lower.
We provide a proof in the supplementary showing that
$c_i$ takes a high value only when predictions are consistent and all with
low entropy.

\noindent \textbf{ConfNet.}
%
%Although aggregating predictions from multiple views can produce a more precise confidence estimation,
%the improvement is limited as it is closely coupled with the local predictor.\todo{a more convincing reason}
% another possible reason:
% it only exploits the confident samples
%In this stage, we regard confidence estimation as a ranking problem.
%
Among the initial confidence estimated from previous stage,
the most confident samples are most likely to be genuine members
while the most unconfident samples are most likely to be outliers,
which can be regarded as positive samples and negative samples, respectively.

ConfNet is introduced to learn from the ``discovered'' genuine members and outliers.
It aims to output a probability value for each vertex $v$ to indicate
how likely it is a genuine member instead of an outlier.
Similar to the local predictor, we implement ConfNet as a graph convolutional
network, following Eq.~\eqref{eq:gcn}.
Given a percentage $\eta$ and sampled graph patches,
we take the top-$\eta$ confident vertices as the positive samples and
the top-$\eta$ unconfident vertices as the negative ones.
Then we train the ConfNet using the vertex-wise binary cross-entropy as the loss function.
The final confidence of a vertex is estimated as the average of
multi-view confidence and the predicted confidence from the learned ConfNet.

\subsection{Training of Local Predictor}

Here we introduce how to train the local predictor.
The training samples consist of graph patches with at least one labeled vertex.
Instead of selecting graph patches consecutively during propagation,
we sample a set of graph patches parallel for training.
The sampling of graph patches follows the same principle, \ie,
selecting those with high confidence.
Based on the sampled subgraphs, the local predictor predicts labels for
all labeled vertices on sampled subgraphs.
The cross-entropy error between predictions and ground-truth is then
minimized over all labeled data to optimize the local predictor.
%The local predicator is then trained
%with the objective to minimize the \emph{cross-entropy} between the predictions
%and the labels over all labeled vertices on the patch.

%The training involves 3 steps.
%(1) We apply Alg.~1 to sample subgraphs.
%(2) The local predictor predicts labels for labeled vertices on sampled
%subgraphs.
%(3) The cross-entropy error between predictions and ground-truth is minimized
%over all labeled data to optimize the local predictor.

%As the propagation proceeds,
%the high-confidence vertex set $\cS$ will gradually expand with highly
%confident samples added in.
%Therefore, the propagation and the training of local predictor can be
%iteratively
%conducted to obtain further performance gain.
%Specifically, we first train the local predictor with labeled samples.
%Then we apply our propagation algorithm with the trained predictor.
%At the end of the propagation, we fine-tune the local predictor with
%expanded confidence set $\cS$ and conduct propagation again.
% !TEX root = ../main.tex

\section{Experiments}
\subsection{Experimental Settings}

\noindent \textbf{Dataset.}
We conduct our experiments on two real-world datasets, namely,
ImageNet~\cite{deng2009imagenet} and Ms-Celeb-1M~\cite{guo2016ms}.
ImageNet compromises $1M$ images from $1,000$ classes, which is the
most widely used image classification dataset.
Ms-Celeb-1M is a large-scale face recognition
dataset consisting of $100K$ identities, and each identity has about $100$
facial images.
Transductive learning in vision tasks considers a practical setting that
obtains a pretrained model but its training data are unavailable.
Given only the pretrained model and another unlabeled set with limited
labeled data,
it aims to predict labels for the unlabeled set.
We simulate this setting with the following steps:
(1) We randomly sample $10\%$ data from ImageNet to train the feature
extractor $\cF$.
(2) We use $\cF$ to extract features for the rest $90\%$ samples to construct $\cD_{all}$.
(3) We randomly sample 10 classes from $\cD_{all}$ as $\cD$, and
randomly split $1\%$ data from $\cD$ as the labeled set $\cD_l$.
(4) With a noise ratio $\rho$, we construct the outlier set
$\cD_o$ by randomly sampling data from $\cD_{all} \setminus \cD$.
(5) $\cD_u$ is a union set of $\cD \setminus \cD_l$ and $\cD_o$.
Experiments on Ms-Celeb-1M follow the same setting except sampling 100 classes.
%In particular, we follow the split of ~\cite{yang2019learning} to construct $\cD_l$ and $\cD_u$ in Ms-Celeb-1M.
%but assume that the training labels for the feature extractor are
%unavailable for later propagation.
%
%$10\%$ images in the original training set are split to
%train the feature extractor in both ImageNet and Ms-Celeb-1M.
%We apply the trained feature extractor on the remaining images to
%get feature representation for each image.
%%
%Keeping only 1\% as the labeled samples, we split the dataset into
%a labeled set $\cD_l$ and an unlabeled set $\cD_u$.
%
We sample a small validation set $\cD_v$ with the same size
as $\cD_l$, to determine the outlier threshold $o_{\tau}$.
To evaluate performance on graphs with different noise ratio,
we set the noise ratio $\rho$ to $0\%$, $10\%$, $30\%$ and $50\%$ .

\noindent \textbf{Metrics.}
We assess the performance under the noisy transductive learning.
Given the ground-truth of the unlabeled set, where the ground-truth of
out-of-class outliers is set to $-1$,
transductive learning aims to predict the label of each sample in $\cD_u$,
where the performance is measured by \emph{top-$1$} accuracy.
%
%To evaluate performance on graphs with different scales, we randomly choose
%$10$, $100$ and $1000$ classes from ImageNet, and $100$, $1000$
%and $8000$ classes from Ms-Celeb-1M for experiments. The graph size
%varies from $10K$ vertices to $1M$ vertices.

\noindent \textbf{Implementation Details.}
We take ResNet-50~\cite{he2016deep} as the feature extractor in our experiments.
$K=30$ is used to build the $K$NN affinity graph.
$c_\tau$ is set to 0.9 as the threshold to fix high confident vertices.
$s$ and $\Delta c_\tau$ for generating graph patches is
3000 and 500.
We use SGCs~\cite{wu2019simplifying} for both local predictor and ConfNet.
The depth of SGC is set to 2 and 1 for local predictor and ConfNet, respectively.
The Adam optimizer is used with a start learning rate 0.01 and the
training epoch is set to 200 and 100 for local predictor and ConfNet, respectively.
%
%The experiments are conducted on ES-2640 v3 and TitanXP.

\subsection{Method Comparison}
\label{sec:exp-comp}

We compare the proposed method with a series of transductive baselines.
Since all these methods are not designed for noisy label propagation,
we adapt them to this setting by adopting the same strategy as our method.
Specifically, we first determine the outlier threshold $o_{\tau}$ on a validation set $\cD_v$,
and then take the samples whose confidence below the threshold $o_{\tau}$ as the noisy samples.
The methods are briefly described below.

\noindent\textbf{(1) LP~\cite{zhu2002learning}}
is the most widely used transductive learning approach,
which aggregates the labels from the neighborhoods by linear combination.

\noindent\textbf{(2) GCN~\cite{kipf2016semi}}
is devised to capture complex graph structure,
where each layer consists of a non-linear transformation and an aggregation function.
%To apply GCN on massive graphs, we apply training and inference
%on random selected local sub-graphs. It is referred as GCN (batch) in
%Tab.~\ref{tab:exp-transductive}.

\noindent\textbf{(3) GraphSAGE~\cite{hamilton2017inductive}}
is originally designed for node embedding, which
applies trainable aggregation functions on sampled neighbors.
We adapt it for transductive learning by replacing the unsupervised
loss with the cross-entropy loss.

\noindent\textbf{(4) GAT~\cite{velivckovic2017graph}}
introduces a self-attention mechanism to GCN,
which enables specifying different weights to different nodes in a neighborhood.

\noindent\textbf{(5) FastGCN~\cite{chen2018fastgcn}} addresses the
memory issue of GCN by a sampling scheme. Compared with GraphSAGE,
it saves more memory by sampling vertices rather than neighbors at each layer.

\noindent\textbf{(6) SGC~\cite{wu2019simplifying}} simplifies the
non-linear transformation of GCN, which comprises a linear local
smooth filter followed by a standard linear classifier.

\noindent\textbf{(7) Ours}
incorporates a local predictor and a confidence-based path scheduler.
The two closely coupled components learn to propagate on noisy graphs reliably.
%
%To show the effectiveness of our approach, we report the results by
%propagating across the graph once and using the ConfNet in a post-processing manner.
%\todo{Is the above sentence clear?}

%\noindent\textbf{(8) Ours (multi-training),} the multi-training version
%of the proposed method. %As the propagation proceeds, the labeled set will be
%expanded with confident samples.
%%
%When propagation proceeds for a number of times, we use the expanded
%labeled set with added confident samples to fine-tune the local predictor.\todo{may remove this one}

\begin{table*}[]
\centering
\caption{Performance comparison of transductive methods on noisy affinity graphs.
	GraphSAGE$^\dagger$ denotes using GCN as the aggregation function.
	For both ImageNet and Ms-Celeb-1M, 1\% labeled images are randomly selected as seeds. We randomly select classes and initial seeds for 5 times and report the average results of 5 runs
	(see supplementary for the standard deviation of all experiments)
	}
\begin{tabular}{l|cccc|cccc}
\hline
 & \multicolumn{4}{c|}{ImageNet} & \multicolumn{4}{c}{Ms-Celeb-1M}  \\ \hline
Noise ratio $\rho$ & 0\% & 10\% & 30\% & 50\% & 0\% & 10\% & 30\% & 50\% \\\hline
LP~\cite{zhu2002learning} & 77.74 & 70.51 & 59.47 & 51.43 & 95.13 & 89.01 & 88.31 & 87.19 \\
GCN~\cite{kipf2016semi} & 83.17 & 75.37 & 66.28 & 64.09 & 99.6 & 99.6 & 96.37 & 96.3 \\
GAT~\cite{velivckovic2017graph} & 83.93 & 75.99 & 66.3 & 63.34 &  99.59 & 96.48 & 94.55 & 94.01 \\
% GCN (batch) & x$\pm$y & x$\pm$y & x$\pm$y &x$\pm$y & x$\pm$y & x$\pm$y  \\
GraphSAGE~\cite{hamilton2017inductive} & 82.42 & 73.42 & 63.84 & 59.12 & 99.57 & 95.68 & 92.21 & 91.06 \\
GraphSAGE$^\dagger$~\cite{hamilton2017inductive} & 81.39 & 73.53 & 63.42 & 58.99 & 99.59 & 95.62 & 92.38 & 91.19 \\
FastGCN~\cite{chen2018fastgcn} & 81.34 & 74.08 & 63.79 & 58.81 & 99.62 & 95.6 & 92.08 & 90.83 \\
SGC~\cite{wu2019simplifying} & 84.78 & 76.71 & 67.97 & 65.63 & 99.63 & 97.43 & 96.71 & 96.5 \\ \hline\hline
\textbf{Ours} & \textbf{85.16} & \textbf{76.96} & \textbf{69.28} & \textbf{68.25} & \textbf{99.66} & \textbf{97.59} & \textbf{96.93} & \textbf{96.81} \\ \hline
	\end{tabular}
	\label{tab:exp-transductive}
	\vspace{-0.5em}
\end{table*}

\begin{figure*}[h]
\centering
 \begin{minipage}[b]{0.49\textwidth}
 \includegraphics[width=0.8\linewidth]{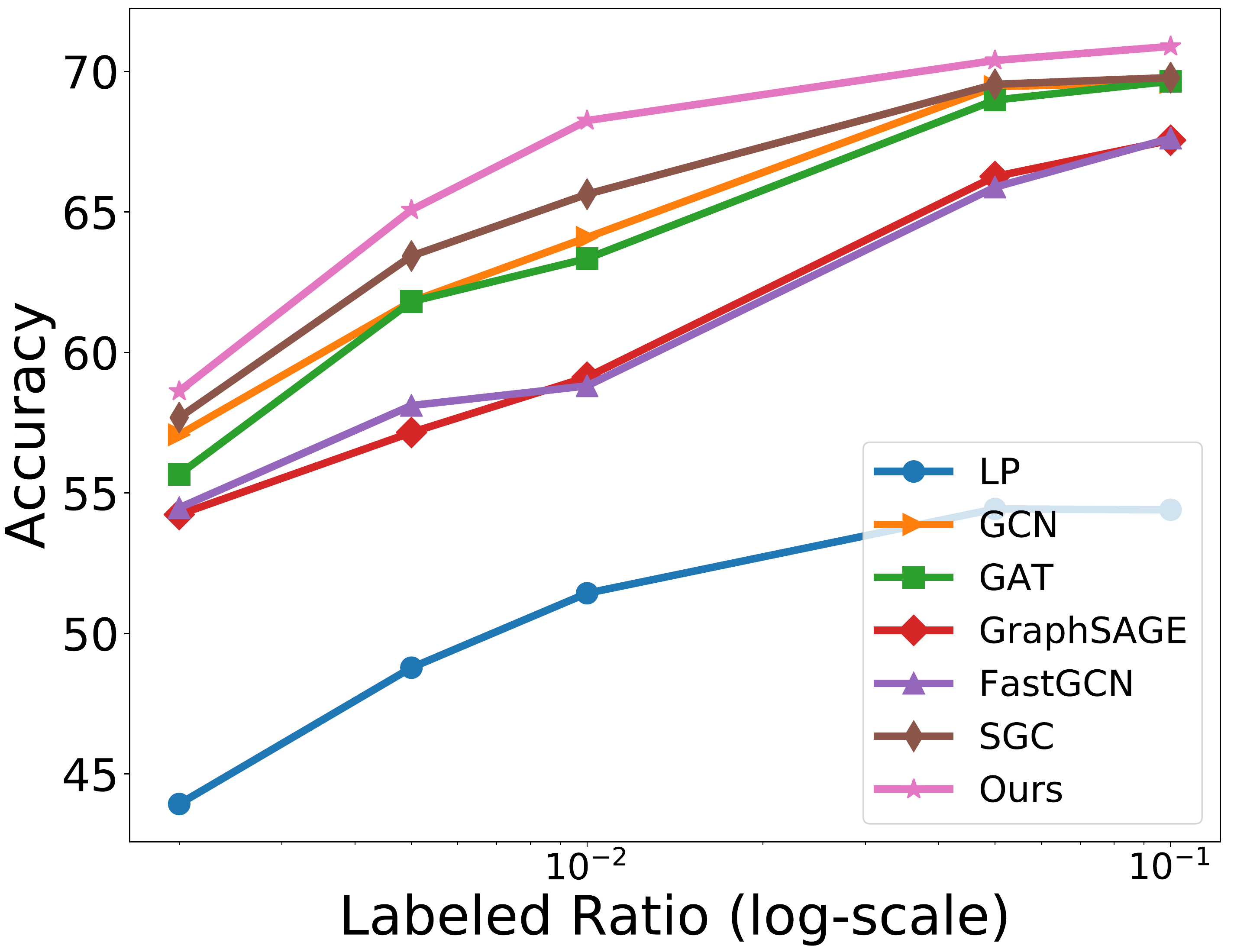}
	\caption{
		Influence of labeled ratio
	}
	\label{fig:exp_labeled_ratio}
 \end{minipage}
 \begin{minipage}[b]{0.49\textwidth}
	\includegraphics[width=0.8\linewidth]{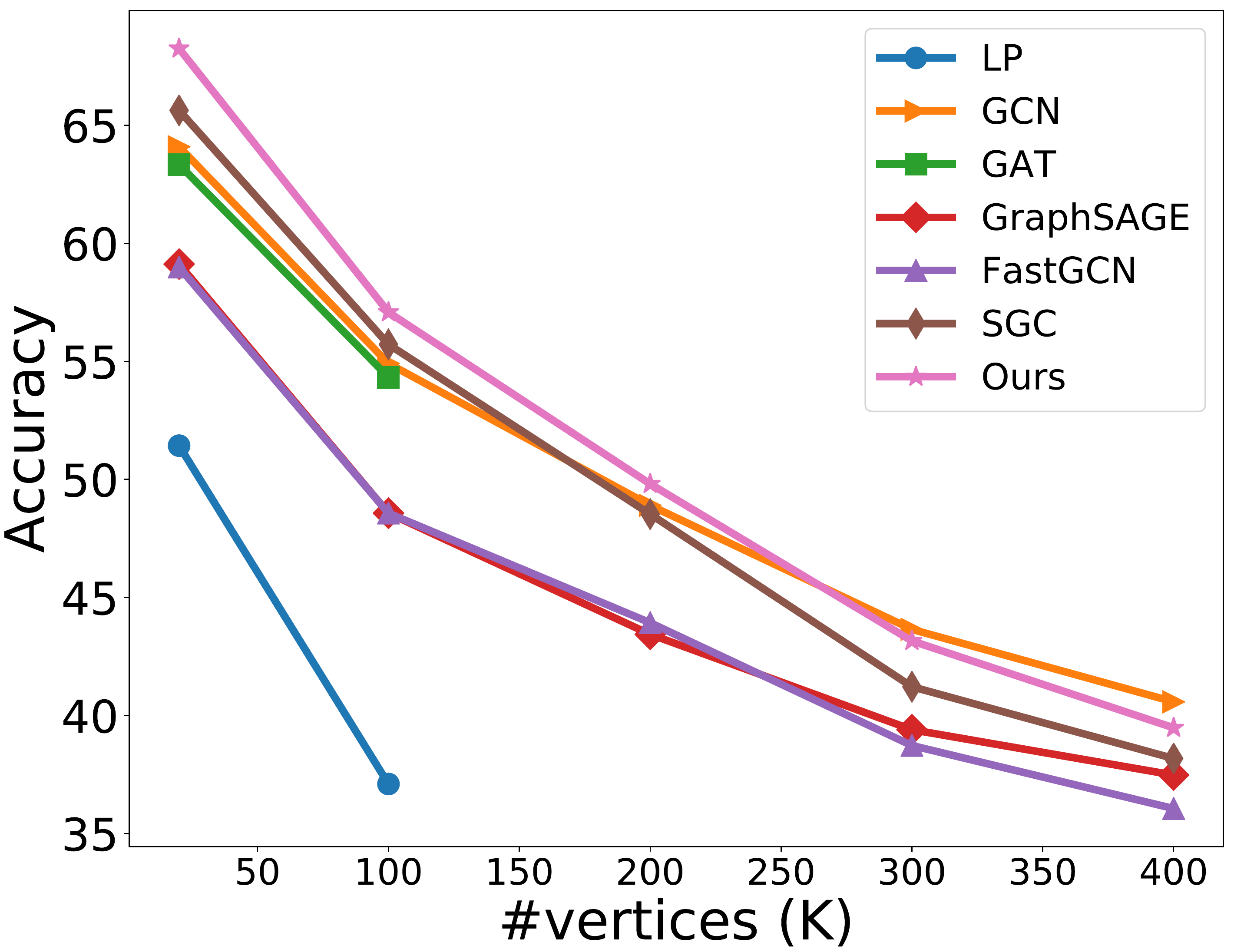}
	\caption{
		Influence of graph size
	}
	\label{fig:exp_graph_scale}
 \end{minipage}
\end{figure*}

\noindent \textbf{Results.}
%
%To better understand the strength and weakness of our approach,
%We compare with existing methods on two datasets, and the results
%are shown in Table.~\ref{tab:exp-transductive}:
Table.~\ref{tab:exp-transductive} shows that:
(1) For LP, the performance is inferior to other learning-based approaches.
(2) GCN shows competitive results under different settings, although
it is not designed for the noisy scenario.
%We use a 2-layer GCN with 256 hidden units at each layer.
%However, since it operates layer-by-layer on the entire graph, it fails
%to scale out to graphs with millions of vertices.
%
%We fix the patch size as $30000$, which takes about 9G GPU memory.
%
%(3) We employ GraphSAGE with GCN aggregation (GraphSAGE$^\dagger$)
%and mean aggregation (GraphSAGE).
%The sample size of GraphSAGE is set to $15$ for both two layers and the batch size is set to $32$.
(3) We employ GraphSAGE with GCN aggregation and mean aggregation.
Although it achieves a higher speedup than GCN, not considering
the confidence of predictions makes the sampling-based method very sensitive to outliers.
(4) Although GAT yields promising results when the graph size is $20K$,
it incurs excessive memory demand when scaling to larger graphs,
as shown in Fig.~\ref{fig:exp_labeled_ratio}.
%
%(5) We sample $6,000$ one-hop neighbors and $1,000$ two-hop neighbors.
Despite FastGCN is efficient, it suffers from the similar problem as GraphSAGE.
(6) SGC, as a simplified version of GCN, achieves competitive results
to GCN and GAT. As it has less training parameters, it may not easily
overfit when the initial seeds are sparse.
Fig.~\ref{fig:exp_labeled_ratio} indicates that the performance of SGC becomes inferior to GCN when the graph size becomes large.
(7) Table.~\ref{tab:exp-transductive} illustrates that the noisy
setting is very challenging, which deteriorates the performance of all algorithms marginally.
%When the proportion of the out-of-class noise is 50\%, the accuracy of
%all previous algorithms drop nearly 20\%.
The proposed method improves the accuracy under all noise ratios,
with more significant improvement as the noise ratio becomes larger.
It not only surpasses the sampling-based approaches by a large margin,
but also outperforms the GNNs with the entire graph as inputs.
Even in the well-learned face manifold, which is less sensitive to out-of-sample noise,
our method still reduces the error rate from 3.5\% to 3.19\%.
Note that the proposed method can be easily extended to the iterative scheme by using self-training~\cite{rosenberg2005semi}.
As it can effectively estimate confidence,
applying it iteratively can potentially lead to better results.

%Since it considers reliability of predictions during propagation, it may
%reduce the influence of noisy predictions.
%
%(8) We iteratively apply training and propagation for $3$ times in our
%multi-training setting,
%which receives further performance gain.
%%
%Since our method considers the reliability of predictions during propagation,
%it could reduce the influence of noisy predictions in the propagation. We
%provide visualizations of local sub-graphs in supplementary to demonstrate how
%our method deals with unconfident samples.

%\begin{figure}[t]
%	\centering
%	\includegraphics[width=0.7\linewidth]{figures/exp_runtime}
%	\caption{
%		Accuracy \emph{vs} Runtime. $c_\tau$ is set to
%		$0, 0.6, 0.8, 0.95$ for each point. Note that x-axis is in \emph{log-scale}.
%	}
%	\label{fig:exp-runtime}
%	\vspace{-1em}
%\end{figure}

%\begin{figure}[t]
%	\centering
%	\includegraphics[width=0.8\linewidth]{figures/exp_labeled}
%	\caption{
%		Influence of labeled ratio and graph size. (x-axis in \emph{log-scale})
%		\todo{replace with ``labeled ratio study'' and ``graph size''}
%	}
%	\label{fig:exp-ratio}
%	\vspace{-1em}
%\end{figure}

\noindent \textbf{Labeled Ratio.}
When the noise ratio $\rho$ is $50\%$, we study the influence of
different labeled ratios: $0.2\%, 0.5\%, 1\%, 5\%$ and $10\%$.
Fig.~\ref{fig:exp_labeled_ratio} shows that our method consistently outperforms
other methods under all labeled ratios.
When the initial seeds are very sparse, it becomes more challenging
for both label propagation and confidence estimation.
As our method learns from the discovered confident and unconfident samples along with the propagation,
our method still performs well when there are a few initial seeds.

%
%When the ratio of labeled data increases, the task becomes less challenging
%and the gap between different algorithms decreases.
%
%When labeled data is very sparse, our method improves accuracy by a significant
%margin.
%%
%It indicates the importance of including confidence under noisy setting,
%especially when initial information is limited.

\noindent \textbf{Graph Scale.}
The local update design makes the proposed method capable of scaling to large-scale graphs.
As Fig.~\ref{fig:exp_graph_scale} illustrates,
LP suffers a severe performance drop when the graph size increases.
GAT exceeds the memory limits when the number of vertices is beyond $100K$.
Two sampling-based methods, GraphSAGE and FastGCN,
are inferior to their counterparts operating on the entire graph.
Although our method also operates on subgraphs, the reliable strategy
enables it to perform well on noisy graphs under different scales.
Note that when the graph size is $400K$, GCN performs better than ours.
As we adopt SGC as the local predictor in our experiments,
without non-linear transformation may limit its capability when graph scale is large.
In real practice, we have the flexibility to select different local predictors according to the graph scale.

\begin{table*}
 \begin{minipage}[b]{0.45\textwidth}
  \centering\small
	\caption{Comparison on local predictors and confidences.
	ConfNet$^\dagger$ computes confidence as the average confidence
	from Multi-view and ConfNet
	}
	\begin{tabular}{l|c|c|c}
		\hline
		Confidence & GAT & GCN & SGC \\ \hline
		Random & 63.16 & 62.62 & 64.84 \\
		Multi-view & 64.11 & 63.84 & 65.81 \\
		ConfNet & 64.83 & 63.93 & 67.79 \\
		ConfNet$^\dagger$ & 65.95 & 65.21 & 68.25 \\ \hline
		GT & 83.17 & 83.93 & 84.78 \\ \hline
	\end{tabular}
	\label{tab:exp-pred-conf}
 \end{minipage}
 \hfill
 \begin{minipage}[b]{0.53\textwidth}
  \centering
	\caption{Comparison on different source of initial confidence.
	FNR denotes \emph{false noise ratio} of positive samples and
	TNR denotes \emph{true noise ratio} of negative samples}
	\begin{tabular}{l|c|c|c|c}
		\hline
		Initial Confidence & Num & FNR & TNR & Acc \\ \hline
		SGC($\eta$=0.05) & 973 & 3.8\% & 66\% & 67.44 \\
		Multi-view($\eta$=0.01) & 194 & 1.6\% & 70\% & 66.78 \\
		Multi-view($\eta$=0.05) & 973 & 3.2\% & 65\% & 68.79 \\
		Multi-view($\eta$=0.1) & 1947 & 4.1\% & 63\% & 67.81 \\ \hline
		GT($\eta$=0.05) & 973 & 0\% & 100\% & 76.29\\ \hline
	\end{tabular}
	\label{tab:exp-conf-net}
 \end{minipage}
\end{table*}

\subsection{Ablation Study}
\label{sec:ablation}

We adopt a setting on ImageNet, where the labeled ratio is 1\% and the noise ratio is 50\%,
to study some important designs in our framework.

\noindent \textbf{Local predictor.}
In our framework, the local predictor can be flexibly replaced
with different graph-based algorithms.
%
%Assuming different local predictors share the same path scheduler,
We compare the effectiveness of three learnable local predictors,
namely GAT, GCN, and SGC.
%
%Given a graph patch, Linear local predictor updates the label of each vertex
%by computing linear weighted combination of its neighbors' labels.
%
All three methods take the vertex features as input,
and predict labels for unlabeled vertices.
Comparing different columns in Table.~\ref{tab:exp-pred-conf},
all three local predictors outperforms LP (see Table.~\ref{tab:exp-transductive}) significantly,
even using random confidence.
The results demonstrate the advantage of learning-based approaches
in handling complex graph structure.
%
%On the other hand, GAT does not show much advantage in prediction accuracy,
%while having a complexity that is quadratic in the number of vertices.
%
%\todo{comparison to LP is somehow weak? any other better analysis?}

\noindent \textbf{Path scheduler.}
As shown in different rows in Table.~\ref{tab:exp-pred-conf}, we study confidence choices with different local predictors.
(1) \emph{Random} refers to using random score between $0$ and $1$ as the
confidence, which severely impairs the performance.
(2) \emph{Multi-view} denotes our first stage confidence estimation,
\ie, aggregating predictions from multiple graph patches, which provides a good initial confidence.
(3) \emph{ConfNet} indicates using the confidence predicted from ConfNet.
Compared to Multi-view, the significant performance gain demonstrates
the effectiveness of ConfNet.
(4) \emph{ConfNet$^\dagger$} is the ultimate confidence in our approach.
It further increases the performance by averaging confidence from two previous stages,
which shows that the confidence from Multi-view and ConfNet may be complementary to some extent.
(5) \emph{GT (Ground-truth)} denotes knowing all outliers in advance,
which corresponds to the setting that noise ratio is 0 in Table.~\ref{tab:exp-transductive}.
It indicates that the performance can be greatly boosted if identifying
all outliers correctly.

%\begin{figure}[t]
%	\centering
%	\includegraphics[width=0.8\linewidth]{figures/exp_confidence}
%	\caption{Comparison on Reliability Diagrams~\cite{guo2017calibration}.
%		The difference between accuracy and confidence for a given bin
%		represents the calibration gap (red bars).
%	}
%	\label{fig:exp_confidence}
%	\vspace{-1em}
%\end{figure}
%
%\begin{figure}[t]
%	\centering
%	\includegraphics[width=0.8\linewidth]{figures/exp_classes}
%	\caption{Mean accuracy \emph{vs} Mean confidence over classes on ImageNet.
%	}
%	\label{fig:exp_classes}
%	\vspace{-1em}
%\end{figure}

\noindent \textbf{Confidence estimation}
Table.~\ref{tab:exp-conf-net} analyzes the source of initial confidence
for ConfNet training.
$\eta$ denotes the proportion of the most confident and unconfident samples, as defined in Sec.~\ref{sec:method_prop}.
SGC refers to using the prediction probabilities without Multi-view strategy.
It shows that:
(1) Comparison between SGC ($\eta$=0.05) and Multi-view ($\eta$=0.05) indicates that ConfNet is affected by the quality of initial confidence set.
As Multi-view gives more precise confidence estimation,
it provides more reliable samples for ConfNet training, leading to a better performance.
(2) Comparison between GT ($\eta$=0.05) and Multi-view ($\eta$=0.05) further indicates that training on a reliable initial confidence set
is a crucial design.
(3) Comparison between Multi-view with three different $\eta$ shows that choosing a proper proportion is important to ConfNet training.
When $\eta$ is small, although the positive and negative samples are more pure, training on a few samples impairs the final accuracy.
When $\eta$ is large, the introduction of noise in both positive and negative samples lead to the limited performance gain.

\begin{figure}[t]
	\centering
	\includegraphics[width=0.95\linewidth]{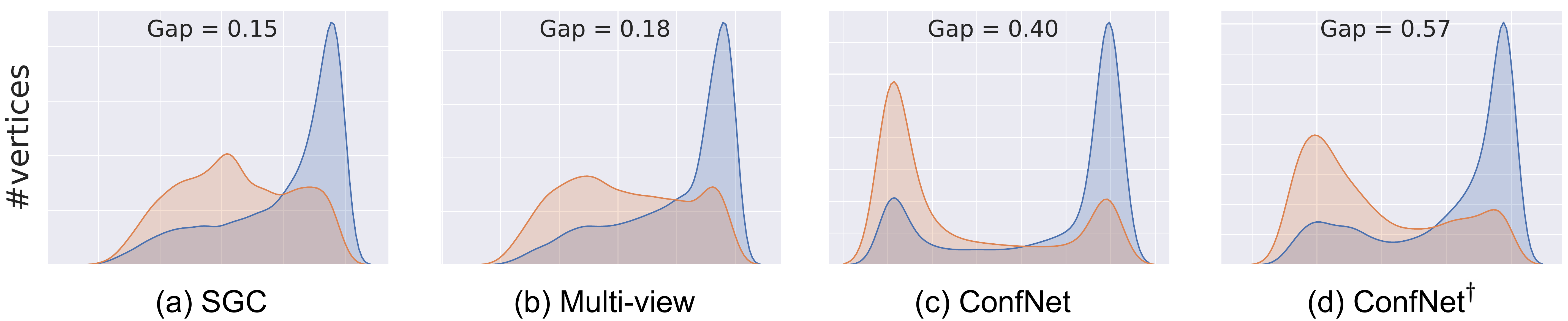}
	\caption{
		Confidence distribution of outliers and genuine members.
		Orange represents the out-of-class noisy samples, while blue
		denotes the in-class unlabeled ones.
		Gap is computed as the difference between the mean of two distributions.
		It indicates that the proposed confidence estimation approach
		can enlarge the confidence gap between outliers and genuine
		ones, which is the key to our performance gain
	}
	\label{fig:exp_conf_dist}
	%\vspace{-1.5em}
\end{figure}

\begin{figure*}[h]
 %\vspace*{-0.2in}
 \begin{minipage}[b]{0.49\textwidth}
  \centering
	\includegraphics[width=0.8\linewidth]{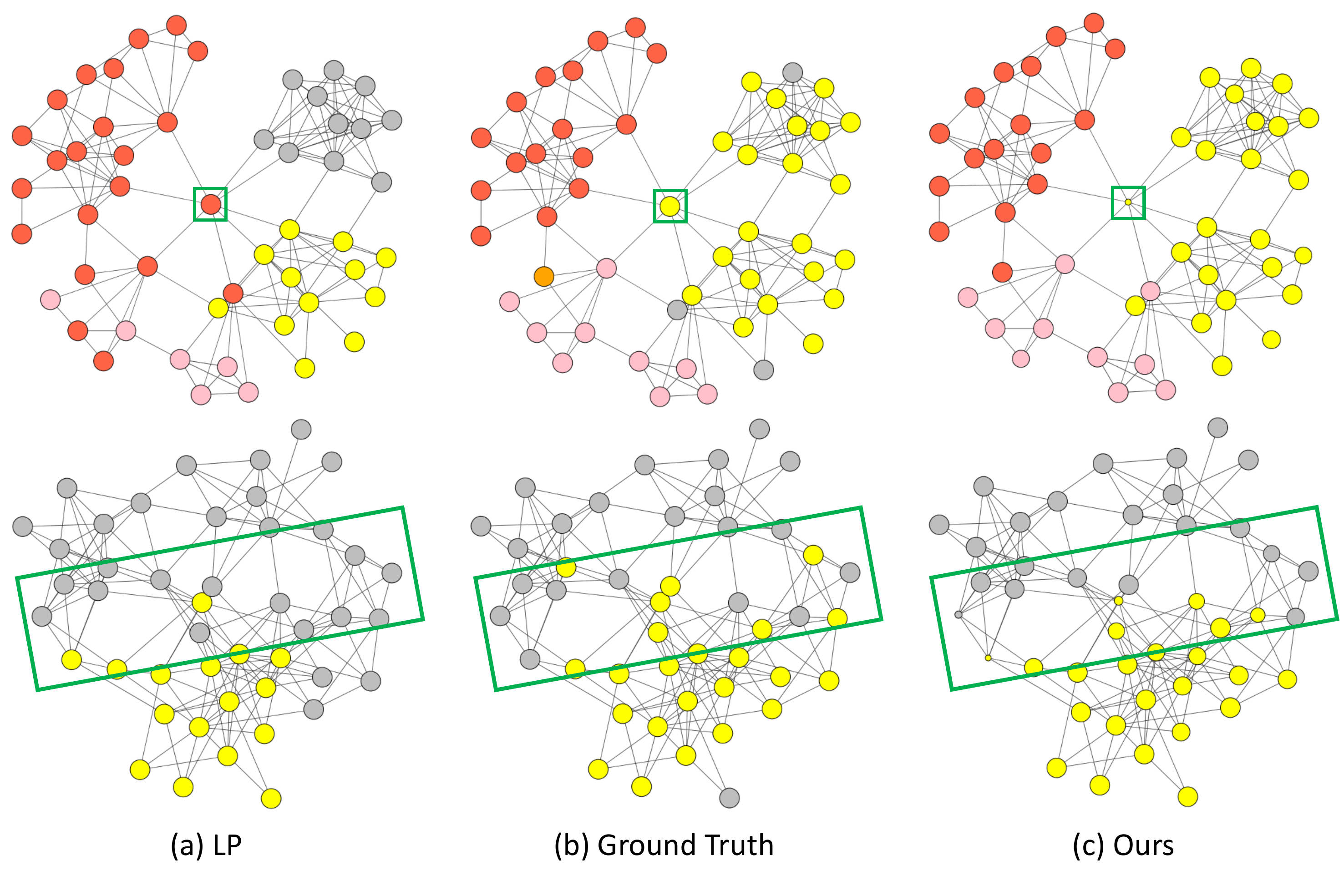}
	\caption{
		%Qualitative Comparison.
		Two graph patches with predictions from ImageNet, where
		different colors represent different classes
	}
	\label{fig:exp_patches}
 \end{minipage}
 \hfill
 \begin{minipage}[b]{0.49\textwidth}
  \centering
	\includegraphics[width=0.82\linewidth]{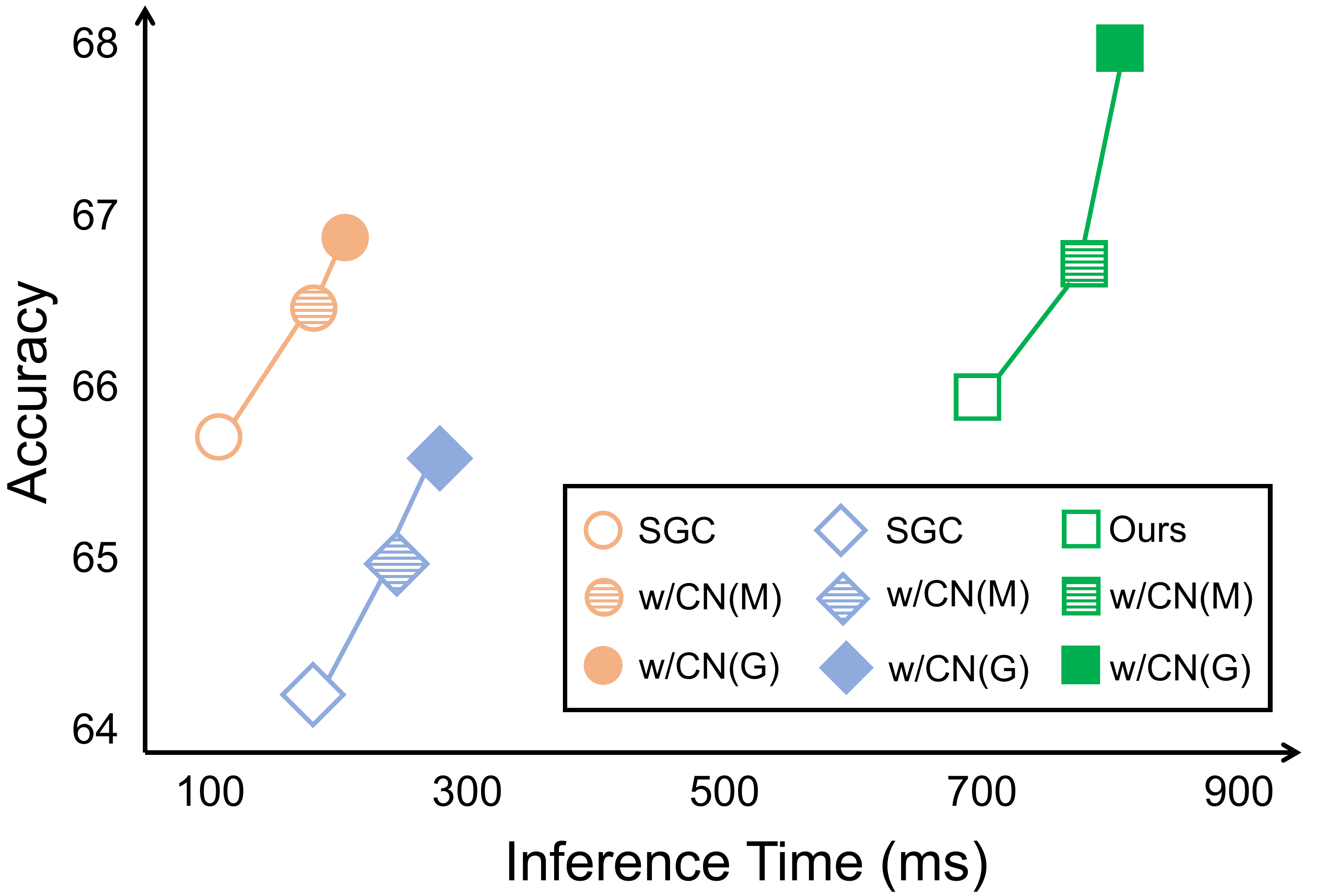}
	\caption{
		Apply ConfNet to different GNNs.
		CN(M) denotes ConfNet using MLP and CN(G) denotes ConfNet using GCN
	}
	\label{fig:confnet}
 \end{minipage}
 %\vspace*{-0.2in}
\end{figure*}

From another perspective,
Fig.~\ref{fig:exp_conf_dist} illustrates that the success of Multi-view and ConfNet is
mainly due to altering the confidence distribution, where the gap between outliers and genuine members is enlarged and thus outliers can be identified more easily.
Fig.~\ref{fig:confnet} shows that using ConfNet as a post-processing module
in previous methods can also improve their capability of identifying outliers,
leading to a significant accuracy gain with limited computational budget.

%\begin{figure}[t]
%	\centering
%	\includegraphics[width=0.8\linewidth]{figures/exp_vt_num}
%	\caption{
%		\#samples \textit{vs.} visited times
%	}
%	\label{fig:vt_num}
%	\vspace{-1em}
%\end{figure}
%
%\begin{figure}[t]
%	\centering
%	\includegraphics[width=0.8\linewidth]{figures/exp_patches}
%\caption{Two graph patches with predictions from ImageNet, where different
%colors represent different classes. }
%	\label{fig:exp_patches}
%	\vspace{-1em}
%\end{figure}

%\vspace{-11pt}
\subsection{Further Analysis}

\noindent \textbf{Efficiency of Path Extraction.}
%
%We use visited time of vertices to measure the efficiency of patch extraction.
%
We refer to \emph{visited times} of a vertex as the number of patches it belongs to.
We conduct experiments on ImageNet with $10K$ vertices
with $c_\tau = 0.9, \Delta c_\tau = 500$ and $s = 3000$.
When propagating $100$ iterations,
the average visited times of vertices are about $6$.
Most samples are visited $2$ times and only a
very few samples are visited more than $10$ times.
%
%As the number of visited times increases, the ratio of incorrect samples over
%correct ones increases, indicating that regions with low confidence will
%be visited more times than those with high confidence.
%
%Besides, at the end of the propagation, the high-confidence set
%contains about $45\%$ of all the samples, while only $0.1\%$ samples has been
%selected as the start vertex. Since our patch extraction considers expected
%confidence gain, it will ignore the start vertices which reside in the local
%patch that already has very high confidence, which shows the effectiveness of
%our approach.

\noindent \textbf{Conservative Prediction on Hard Cases.}
Except the out-of-sample noise, we also visualize the low confident
samples when noise ratio is 0.
%As shown in Fig.~\ref{fig:exp_patches},
%many real-world data are noisy with blurry boundaries between different groups.
%
%For our method, a small size of
%vertex indicates a low confident prediction.
%The first row shows
%the difference between LP and our method when dealing with a hard case (inside the green box).
As Fig.~\ref{fig:exp_patches} shows,
when dealing with a hard case (the green box in the first row),
our method gives the right prediction with very low confidence (small size of vertices)
while LP gives a wrong prediction and misleads the predictions of downstream vertices.
The second row shows that our confidence can identify
inter-class boundaries, and remain conservative to these predictions,
as highlighted in the green box.

%\vspace{-3pt}
\begin{table}[ht]
    %\vspace*{-0.15in}
    \centering
     \caption{Two applications of our method in vision tasks.
	(a) We use the estimated confidence as indicators in active learning.
	(b) We apply the predicted labels to face recognition training in an inductive manner.
	(see supplementary for more details)
	}
	\label{tab:exp-applications}
    \begin{minipage}{0.48\textwidth}
    %\begin{subtable}{1.0\textwidth}
    \centering
    \scalebox{1.0}{\begin{tabular}{c|c|c|c|c}
        \hline
		Labeled & Baseline & Random & GCN & Ours \\ \hline
		1\% & 65.63 & 65.71 & 66.7 & 68.6 \\ \hline
		%10\% &  &  &  & \\ \hline
    \end{tabular}} \\
    (a)
    %\caption{}
    %\label{tbl:trans3}
    %\end{subtable}
    \end{minipage}
    \begin{minipage}{0.48\textwidth}
    %\begin{subtable}{1.0\textwidth}
    \centering
    \scalebox{1.0}{\begin{tabular}{l|c|c|c|c}
        \hline
		Test Protocol & Baseline & CDP & GCN & Ours \\ \hline
		%F1-score & - &  &  &  \\ \hline
		MegaFace~\cite{kemelmacher2016megaface} & 58.21 & 59.15 & 59.33 & 60.02 \\ \hline
    \end{tabular}} \\
    (b)
    %\caption{}
    %\label{tbl:highlow}
    %\end{subtable}
    \end{minipage}
    %\vspace*{-0.2in}
\end{table}

%\vspace{-21pt}
\subsection{Applications}

\noindent \textbf{Active Learning.}
Active learning desires an effective indicator to select representative unlabeled samples.
Table.~\ref{tab:exp-applications}(a) shows that our estimated confidence outperforms two widely used indicators.
Specifically,
the first one \emph{randomly} selects unlabeled samples for annotation,
while the second one applies a trained \emph{GCN} to unlabeled samples and select those with large predicted entropy.
\emph{Baseline} refers to the accuracy before annotation.
%
%All methods annotate as much unlabeled data as the initial labeled data.
%and report the accuracy of using both initial labeled data and extra annotated data for propagation.
%
The result shows that our method brings larger accuracy gain by
annotating the same number of unlabeled data.
%under both $1\%$ and $10\%$ settings.
 
%\emph{GCN} refers to the idea that selecting samples with largest entropy from a trained GCN
%%
%To be specific, we use different indicators to select samples for
%annotation.
%We compare with three indicators in Table.~\ref{tab:exp-applications}(a).
%\emph{Baseline} indicates the results before annotation.
%The first one \emph{randomly} selects unlabeled samples randomly for annotation.
%\emph{GCN} selects
%\emph{Ours} selects samples
%We then add the annotated samples to labeled set and apply
%the same algorithm again.
%%
%The number of annotated samples is the same as that of the original labeled
%samples.
%%
%The performance gap between using extra annotated data and baseline
%illustrates the effectiveness of the indicator.
%%
%As shown in Table.~\ref{tab:exp-applications}(a), \emph{Baseline} indicates
%the accuracy of original algorithm.
%\emph{Random} and \emph{Uncalibrated} correspond to indicators defined in Sec.
%\ref{sec:ablation}.
%\emph{Confidence} refers to our proposed method.
%%Three left columns correspond to three different indicators, as defined in the
%paper.
%%
%\emph{Random} only brings a slightly performance gain since it may annotate
%a lot of easy samples.
%%
%\emph{Uncalibrated} improves the accuracy through annotating some
%important or hard samples.
%%
%Our calibrated confidence matches the confidence with actual
%accuracy and brings large performance gain by annotating the same
%number of unlabeled data, under both $1\%$ and $10\%$ settings.

\noindent \textbf{Inductive Learning.}
The predicted labels from transductive learning can be used as ``pseudo labels'' in inductive learning.
We randomly selects $1K$ person with $120K$ images from Ms-Celeb-1M,
sampling 1\% as the labeled data.
We compare with CDP~\cite{zhan2018consensus} and GCN~\cite{kipf2016semi} for generating ``pseudo labels''.
%CDP~\cite{zhan2018consensus} is a recent face clustering method and
%does not require the initial labeled seeds.
%
Compared to these two methods, Table.~\ref{tab:exp-applications}(b) shows
our method brings larger performance gain on MegaFace~\cite{kemelmacher2016megaface},
which demonstrates that the proposed method generates pseudo labels with higher quality.
% GCN only performs slightly better than the unsupervised CDP.

%\input{sections/extension.tex}
% !TEX root = ../main.tex

%\vspace{-11pt}
\section{Conclusion}
\label{sec:conclusion}

In this paper, we propose a reliable label propagation approach to extend the transductive learning to a practical noisy setting.
The proposed method consists of two learnable components,
namely a GCN-based local predictor and a confidence-based path scheduler.
Experiments on two real-world datasets show that the proposed approach outperforms previous
state-of-the-art methods with reasonable computational cost.
Ablation study shows that exploiting both confident and unconfident
samples is a crucial design in our confidence estimation.
Extending the proposed method to different kinds of noise, such as adversarial noise~\cite{qiu2019semanticadv},
is desired to be explored in the future.

%In this paper,
%we propose an learnable approach for reliable label propagation
%in massive affinity graphs,
%which consists of a patch extractor, a local predictor and
%a confidence-driven path scheduler.
%We experimentally and theoretically show that using multiple views is
%a crucial design in confidence estimation.
%Experiments on two large-scale datasets show that our confidence guided approach outperforms previous
%state-of-the-art methods with reasonable computational cost.
%The performance gain is more significant as the graph size increases,
%which shows its effectiveness on massive real-world data.
%
%The calibrated confidence estimation in can be useful for other applications, such as active learning.
%\todo{the performance is far from satisfactory under this practical setting, namely, unlabeled data may not necessarily share the same class with the initial seed}
%
%\todo{except the out-of-sample noise in the real-world, the emergence of adversarial samples become a new kind of possible noise source. How to extend our methods to boarder noise scenario remains to be studied in the future. (or in the future, xx is desired to tackle different kinds of noise.}

%\vspace{7pt}
%\noindent \textbf{Acknowledgement} This work is partially supported by the Collaborative Research grant from SenseTime Group (CUHK Agreement No. TS1610626 \& No. TS1712093), the Early Career Scheme (ECS) of Hong Kong (No. 24204215), the General Research Fund (GRF) of Hong Kong (No. 14236516, No. 14203518 \& No. 14241716), and Singapore MOE AcRF Tier 1 (M4012082.020).
\noindent \textbf{Acknowledgment} This work is partially supported by the SenseTime Collaborative Grant on Large-scale Multi-modality Analysis (CUHK Agreement No. TS1610626 \& No. TS1712093), the General Research Fund (GRF) of Hong Kong (No. 14203518 \& No. 14205719).

\bibliographystyle{splncs04}
\bibliography{egbib}
\end{document}

\pagestyle{headings}
\mainmatter
\def\ECCVSubNumber{2376}  % Insert your submission number here

\title{Learn to Propagate Reliably on \\ Noisy Affinity Graphs \\ (Supplementary Materials)}
% Replace with your title

% INITIAL SUBMISSION
\begin{comment}
\titlerunning{ECCV-20 submission ID \ECCVSubNumber}
\authorrunning{ECCV-20 submission ID \ECCVSubNumber}
\author{Anonymous ECCV submission}
\institute{Paper ID \ECCVSubNumber}
\end{comment}

\author{Lei Yang\inst{1}\orcidID{0000-0002-0571-5924} \and
Qingqiu Huang\inst{1}\orcidID{0000-0002-6467-1634} \and \\
Huaiyi Huang\inst{1}\orcidID{0000-0003-1548-2498} \and
Linning Xu\inst{2}\orcidID{0000-0003-1026-2410} \and
Dahua Lin\inst{1}\orcidID{0000-0002-8865-7896}}
\authorrunning{L. Yang and Q. Huang and H. Huang and L. Xu and D. Lin}
% First names are abbreviated in the running head.
% If there are more than two authors, 'et al.' is used.
%
\institute{The Chinese University of Hong Kong \and
The Chinese University of Hong Kong, Shenzhen \\
%\email{lncs@springer.com}\\
%\url{http://www.springer.com/gp/computer-science/lncs} \and
%ABC Institute, Rupert-Karls-University Heidelberg, Heidelberg, Germany\\
\email{\{yl016,hq016,hh016,dhlin\}@ie.cuhk.edu.hk,linningxu@link.cuhk.edu.cn}}

\maketitle

%%%%%%%%% BODY TEXT

\section{Prove properties of multi-view confidence.}
Let $\vp_1, \vp_2, ..., \vp_n$ be the $n$ predictions obtained from $n$ \emph{views} for a vertex $v$,
where $\vp_i = (p_{i1}, p_{i2}, ..., p_{im})$, $m$ is the number of classes, and $\sum_{j=1}^m p_{ij}= 1$,
$p_{ij} \in [0, 1]$.
The final prediction of $v$ is defined as
\begin{equation}
\vp = (\overline{p}_{1}, \overline{p}_{2}, ..., \overline{p}_{m}),
\label{eq:pred}
\end{equation}
where $\overline{p}_{j} = \frac{1}{n} \sum_{i=1}^n p_{ij}$, $\sum_{j=1}^m \overline{p}_{j} = 1$.
Recall the confidence of $v$ is defined as
\begin{equation}
c = \max_j \overline{p}_j,
\end{equation}
Our hypothesis is that $c$ takes a high value only when predictions are
consistent and all with low entropy.

First, we prove that a high value of $c$ implies predictions are consistent.
Consider $n$ prediction values on dimension $j$, \ie, $\{p_{1j}, p_{2j}, ...,
p_{nj}\}$.
We use variance $\sigma_j^2$ to describe the consistency of predictions.
The smaller $\sigma_j^2$, the higher consistency of predictions, and vice versa.
$\sigma_j^2$ can be represented as
\begin{equation}
\sigma_j^2 = \frac{1}{n} \sum_{i=1}^n p_{ij}^2 - (\frac{1}{n} \sum_{i=1}^n
p_{ij})^2
= \frac{1}{n}\sum_{i=1}^n p_{ij}^2 - \overline{p}_j^2
\label{eq:sigma}
\end{equation}
From Eq.~\ref{eq:sigma} and $p_{ij} \in [0, 1]$, we have,
\begin{equation}
\begin{aligned}
\overline{p}_j &= \sqrt{\frac{1}{n} \sum_{i=1}^n p_{ij}^2 - \sigma_j^2}
\leq \sqrt{1 - \sigma_j^2} \\
\overline{p}_j &= \frac{1}{n} \sum_{i=1}^n p_{ij} \geq \frac{1}{n} \sum_{i=1}^n
p_{ij}^2
= \overline{p}_j^2 + \sigma_j^2 \geq \sigma_j^2
\end{aligned}
\end{equation}
Thus, c is bounded by
\begin{equation}
\sigma_k^2 \leq c \leq \sqrt{1 - \sigma_k^2},
\end{equation}
where $k = \argmax_j \overline{p}_j$.
Hence, the bound of $c$ depends on $\sigma_k$.
When $\sigma_k$ is $0$, the lower bound of $c$ is $0$ and the upper bound of
$c$ is 1.
As $\sigma_k$ increases, the lower bound of $c$ increases and the upper bound
of $c$ decreases.
Therefore, one necessary condition for $c$ to take a high value is the small
variance of the prediction values, which indicates the high consistency of
predictions.

Second, we prove that a high value of $c$ implies predictions have low entropy.
Since we have proved that a high value of $c$ implies predictions are
consistent,
we only need to consider the entropy of final prediction (Eq.~\ref{eq:pred}).
The normalized entropy $E$ of $\vp$ can be written as
\begin{equation}
E = -\sum_j \overline{p}_j \log_m \overline{p}_j = - c \log_m c + E',
\label{eq:entropy}
\end{equation}
where $E'$ includes $m - 1$ values except
the maximum one.
To normalize the maximum value of entropy to $1$, we use $m$ as the
base of log.
For simplicity, we do not mention it in the proof below.
By Jensen's inequality, we have
\begin{equation}
\begin{aligned}
E' &= -\sum_{j \neq k} \overline{p}_j \log \overline{p}_j \\
&\leq - (\sum_{j \neq k} \overline{p}_j) \log \frac{\sum_{j \neq
k}\overline{p}_j}{m-1}
= -(1 - c)\log \frac{1 - c}{m - 1}
\end{aligned}
\end{equation}
From (6)(7), we have
\begin{equation}
E \leq - c \log c - (1 - c)\log \frac{1 - c}{m - 1},
\end{equation}
Let $f(c) = - c \log c - (1 - c)\log \frac{1 - c}{m - 1}$. Taking the
derivative of $f(c)$, we have
\begin{equation}
\frac{\partial f(c)}{\partial c} = \log \frac{1 - c}{c (m - 1)} \leq 0,
\label{eq:derivative}
\end{equation}
where $\frac{1}{m} \leq c \leq 1$, as $c$ is the maximum value.
Since Eq.~\ref{eq:derivative} is consistently below $0$, it
illustrates that $f(c)$ monotonically decreases when $c \in [\frac{1}{m}, 1]$.
When $E$ takes a small value, it implies that the lower bound of $f(c)$
decreases and thus the upper bound of $c$ increases;
When $E$ takes a high value, it implies that the lower bound of $f(c)$
increases and thus the upper bound of $c$ decreases.
Therefore, another necessary condition for $c$ to take a high value is
the small lower bound of $f(c)$, which indicates $\vp$ has small entropy.

From previous discussions, we prove that $c$ takes a high value only
when predictions are consistent and all with low entropy.

\section{Experimental results with standard deviation}

As introduced in Sec.~4.1 in the main text, we repeat our experiments by 5 runs.
For each run, we first randomly select 10 classes from $\cD_{all}$ and randomly split 1\% as the labeled initial seeds.
Then, we randomly select $\rho$ percent of images not belong to the 10 classes as the noise.
As shown in the two tables below,
we observe that the randomly selected 10 classes result in a large standard deviation among different runs,
but the proposed method consistently outperforms previous approaches and thus achieving higher mean accuracy.

\begin{table*}[t]
\centering
\caption{Performance comparison of transductive methods on noisy affinity graphs in ImageNet.
	We randomly select classes and initial seeds for 5 times and report the average results of 5 runs
	}
\begin{tabular}{l|cccc}
\hline
 & \multicolumn{4}{c}{ImageNet} \\ \hline
Noise ratio $\rho$ & 0\% & 10\% & 30\% & 50\% \\\hline
LP~\cite{zhu2002learning} & 77.74$\pm$2.65 & 70.51$\pm$2.27 & 59.47$\pm$1.8 & 51.43$\pm$1.48 \\
GCN~\cite{kipf2016semi} & 83.17$\pm$3.65 & 75.37$\pm$3.12 & 66.28$\pm$2.87 & 64.09$\pm$2.69 \\
GAT~\cite{velivckovic2017graph} & 83.93$\pm$3.36 & 75.99$\pm$2.56 & 66.3$\pm$3.12 & 63.34$\pm$2.06 \\
% GCN (batch) & x$\pm$y & x$\pm$y & x$\pm$y &x$\pm$y & x$\pm$y & x$\pm$y  \\
GraphSAGE~\cite{hamilton2017inductive} & 82.42$\pm$1.07 & 73.42$\pm$2.88 & 63.84$\pm$3.42 & 59.12$\pm$3.45 \\
GraphSAGE$^\dagger$~\cite{hamilton2017inductive} & 81.39$\pm$3.37 & 73.53$\pm$2.85 & 63.42$\pm$2.82 & 58.99$\pm$3.88 \\
FastGCN~\cite{chen2018fastgcn} & 81.34$\pm$3.84 & 74.08$\pm$3.24 & 63.79$\pm$3.24 & 58.81$\pm$2.76 \\
SGC~\cite{wu2019simplifying} & 84.78$\pm$3.35 & 76.71$\pm$3.0 & 67.97$\pm$2.51 & 65.63$\pm$2.58 \\ \hline\hline
\textbf{Ours} & \textbf{85.16$\pm$3.24} & \textbf{76.96$\pm$2.85} & \textbf{69.28$\pm$2.45} & \textbf{68.25$\pm$1.89} \\ \hline
	\end{tabular}
	\label{tab:exp_imagenet}
\end{table*}
\begin{table*}[h]
\centering
\caption{Performance comparison of transductive methods on noisy affinity graphs in Ms-Celeb-1M.
	We randomly select classes and initial seeds for 5 times and report the average results of 5 runs
	}
\begin{tabular}{l|cccc}
\hline & \multicolumn{4}{c}{Ms-Celeb-1M (1\%)}  \\ \hline
Noise ratio $\rho$ & 0\% & 10\% & 30\% & 50\% \\\hline
LP~\cite{zhu2002learning} & 95.13$\pm$1.29 & 89.01$\pm$1.33 & 88.31$\pm$1.17 & 87.19$\pm$0.99 \\
GCN~\cite{kipf2016semi} &  99.6$\pm$0.11 & 99.6$\pm$0.38 & 96.37$\pm$0.42 & 96.3$\pm$0.36 \\
GAT~\cite{velivckovic2017graph} & 99.59$\pm$0.04 & 96.48$\pm$0.51 & 94.24$\pm$0.68 & 94.01$\pm$0.73 \\
% GCN (batch) & x$\pm$y & x$\pm$y & x$\pm$y &x$\pm$y & x$\pm$y & x$\pm$y  \\
GraphSAGE~\cite{hamilton2017inductive} & 99.57$\pm$0.1 & 95.68$\pm$0.28 & 92.21$\pm$0.58 & 91.06$\pm$0.48 \\
GraphSAGE$^\dagger$~\cite{hamilton2017inductive} & 99.59$\pm$0.08 & 95.62$\pm$0.37 & 92.38$\pm$0.4 & 91.19$\pm$0.46 \\
FastGCN~\cite{chen2018fastgcn} & 99.62$\pm$0.07 & 95.6$\pm$0.35 & 92.08$\pm$0.52 & 90.83$\pm$0.81 \\
SGC~\cite{wu2019simplifying} & 99.63$\pm$0.07 & 97.43$\pm$0.35 & 96.71$\pm$0.32 & 96.5$\pm$0.31 \\ \hline\hline
\textbf{Ours} & \textbf{99.66$\pm$0.05} & \textbf{97.59$\pm$0.34} & \textbf{96.93$\pm$0.25} & \textbf{96.81$\pm$0.24} \\ \hline
	\end{tabular}
	\label{tab:exp_msceleb}
\end{table*}

\section{More Details and Analysis}

\noindent \textbf{Details about experimental settings.}
All algorithms use the same affinity graph constructed as follows.
We regard each sample as a vertex and connect it with its $K$ nearest samples.
The edge weight $e_{i, j}$ is the cosine similarity between $v_i$ and
$v_j$ if $(i, j) \in \cE$, otherwise it is zero.
$\cG$ can be alternatively represented as a sparse affinity matrix $\mA$, where the space complexity is $O(NK)$.
The experiments show that the results are not sensitive
to $K$ when varying it from $30$ to $80$ on two datasets.
Therefore, we choose $K=30$ to reduce memory overhead.
For GCN, we use a 2-layer GCN with 256 hidden units at each layer.
For GraphSAGE, the sample size of GraphSAGE is set to $15$ for both two layers and the batch size is set to $32$.
For FastGCN, we sample $6,000$ one-hop neighbors and $1,000$ two-hop neighbors.

\noindent \textbf{Details about efficiency of path extraction.}
%
%We use visited time of vertices to measure the efficiency of patch extraction.
%
We refer to \emph{visited times} of a vertex as the number of patches it belongs to.
We conduct experiments on ImageNet with $10K$ vertices
with $c_\tau = 0.9, \Delta c_\tau = 500$ and $s = 3000$.
When propagating $100$ iterations,
the average visited times of vertices are about $6$.
Most samples are visited $2$ times and only a
very few samples are visited more than $10$ times.
As the number of visited times increases, the ratio of incorrect samples over
correct ones increases, indicating that regions with low confidence will
be visited more times than those with high confidence.

Besides, at the end of the propagation, the high-confidence set
contains about $45\%$ of all the samples, while only $0.1\%$ samples has been
selected as the start vertex. Since our patch extraction considers expected
confidence gain, it will ignore the start vertices which reside in the local
patch that already has very high confidence, which shows the effectiveness of
our approach.

\noindent \textbf{Details about active learning experiments.}
To test the effectiveness of indictors in active learning,
we use three different indicators to select the same number of unlabeled samples for annotation.
All methods annotate as much unlabeled data as the initial labeled data.
The unlabeled \emph{in-class} samples are annotated to their ground-truth classes
and the unlabeled \emph{out-of-class} samples are annotated to -1.
For a fair comparison, we train a standard GCN~\cite{kipf2016semi}
using the previous initial labeled samples in conjunction with the annotated samples,
where the outliers are excluded in the training.
Note that if applying the proposed propagation algorithm,
it can also exploit the annotated out-of-class samples for confidence estimation.

The experimental setting is the same as the other ablation studies in the main text,
where the labeled ratio is 1\% and the noise ratio is 50\%.
The number of annotated samples is also 1\%.
As shown in Table.~4(a) in the main text,
\emph{Baseline} indicates the accuracy of GCN before annotation.
\emph{Random} only brings a slight performance gain as it may
randomly select a lot of easy samples for annotation.
\emph{GCN} improves accuracy through annotating some low confident samples,
which are hard for GCN to recognize in previous training.
Our estimated confidence addresses the noise issue on the unlabeled data,
leading to larger performance gain by annotating the same number of unlabeled data.
% maybe we can show the percentage of pinpointed noise samples here.

Except the out-of-sample noise in the real-world, the emergence of adversarial samples become a new kind of possible noise source~\cite{qiu2019semanticadv}. How to extend our methods to boarder noise scenario remains to be studied in the future.

\noindent \textbf{Details about inductive learning experiments.}
We show that the proposed method can be applied to inductive learning
by providing the ``pseudo labels''.
We compare with two methods for generating ``pseudo labels'' and
apply them to train face recognition model in a supervised manner.
CDP~\cite{zhan2018consensus} is a recent face clustering method and
does not require the labeled seeds.
GCN~\cite{kipf2016semi} leverages the initial labeled seeds and
propagate the labels to the rest unlabeled samples.
Baseline refers to the result before training with pseudo labels.

For the experiment,
we randomly select $1K$ person with $120K$ images from Ms-Celeb-1M and randomly samples 1\% as the labeled data.
To simulate the noisy setting, we then randomly sample $60K$ face images outside the selected $1K$ person.
We use the same pretrained face model from ~\cite{yang2019learning} and regard its performance as the \emph{Baseline}.
As shown in Table.~4(b) in the main text,
without confidence design, GCN only performs slightly better than the unsupervised single version of CDP.
Our method surpasses the previous two methods by providing more accurate pseudo labels.
As the face recognition model is well learned,
it is significant to improve $\sim$2\% on the challenging MegaFace~\cite{kemelmacher2016megaface},
using only $180K$ noisy unlabeled data.
In real practice, instead of discarding the identified noise,
we can further boost the performance by performing
clustering algorithms~\cite{yang2019learning,yang2020learning} on the identified our-of-class samples,
which may discover some new classes for training.

\noindent \textbf{The influence of graph patches with different scales.}
The scale of graph patches affects the efficiency and accuracy of propagation. In our approach, the scale of the graph patch is \emph{dynamic}. Small patches would be extracted at the beginning, since most of the samples are unlabeled and the expected confidence gain is easy to achieve. As the propagation proceeds, the number of confident vertices increases while the average expected confidence gain decreases, the algorithm encourages more aggressive updates over larger patches.
\begin{table}[h]
\centering
\caption{Influence of maximum patch size}
\begin{tabular}{c|c|c|c}
\hline
Maximum patch size ($s$) & Accuracy & Memory & Runtime \\ \hline
100 & 64.63 & 6M & 34s \\
500 & 65.41 & 28M & 111s \\
1000 & 65.64 & 56M & 158s \\
3000 & 65.83 & 168M & 316s \\
5000 & 65.81 & 280M & 564s \\
10000 & 65.92 & 561M & 1077s \\ \hline
\end{tabular}
\end{table}
Although we cannot directly control the scale of each graph patch in our approach, we can change the maximum patch size, which would also show the influence of patch size. We set $\Delta c_\tau = 500$ and vary maximum patch size  from $100$ to $10000$. As shown in the table above, when the maximum size of graph patch is too small ($s=100$), the graph convolutional networks can only leverage a limited number of neighbors for GCN prediction, resulting in the inferior performance. The accuracy increases with the increase of $s$ and it gradually saturates as $s$ increases beyond $3000$. When the maximum size of graph patch is close to the entire graph size ($s=10000$), it introduces a large amount of computational cost and memory overhead but receives slightly performance gain.

\begin{table}[h]
\centering
\caption{Design choices to include confidence into Eq.1}
\begin{tabular}{l|c|c|c}
\hline
Design & Accuracy & Memory & Runtime \\ \hline
(1) & 61.5 & 168M & 316s \\ \hline
(2) & 65.9 & 205M & 386s \\ \hline
Ours & 65.8 & 168M & 316s \\ \hline
\end{tabular}
\end{table}

\noindent \textbf{Design choices to include confidence into Eq.1.}
We investigate two design choices to include confidence into Eq.1. in the main text.
First, we can apply the suggested Hadamard product. As the initial confidence of unlabeled data is small, their node features after multiplication is negligible in GCN's prediction. Ignoring neighbor information may potentially impair the initial predication and the propagation later on.
Second, we can concatenate the confidence distribution $\vp$ to vertex feature $\vx$. Although it does not suffer from initial confidence, the vertex features would be a $N \times (D + M)$ matrix, where $M$ is the number of classes. When $M$ is large, it introduces large computational cost and memory demand.

\bibliographystyle{splncs04}
\bibliography{egbib}